\documentclass[runningheads]{llncs}

\usepackage{eccv}

\usepackage{eccvabbrv}

\usepackage{graphicx}
\usepackage{booktabs}
\usepackage{multirow}

\usepackage[accsupp]{axessibility}  

\usepackage{hyperref}

\usepackage{orcidlink}

\usepackage{xstring}
\usepackage{catchfile}
\let\originalincludegraphics\includegraphics
\renewcommand{\includegraphics}[2][]{%
  \def\originalpath{#2}%
  \StrSubstitute{\originalpath}{(}{_}[\tempA]%
  \StrSubstitute{\tempA}{)}{_}[\tempB]%
  \StrSubstitute{\tempB}{ }{_}[\cleanpath]%
  \originalincludegraphics[#1]{\cleanpath}%
}

\begin{document}

\title{Towards Practical Lossless Neural Compression for LiDAR Point Clouds} 


\author{Pengpeng~Yu\inst{1,2}\textsuperscript{*} \and
Haoran~Li\inst{1,2}\textsuperscript{*} \and
Runqing~Jiang\inst{1,2} \and
Dingquan~Li\inst{2} \and \\
Jing~Wang\inst{2} \and
Liang~Lin\inst{1,2} \and
Yulan~Guo\inst{1,2}\textsuperscript{**}
}

\authorrunning{P.~Yu et al.}

\institute{Sun Yat-sen University, China \and
Pengcheng Laboratory, China\\
\url{guoyulan@sysu.edu.cn} }

\maketitle

\begin{abstract}
LiDAR point clouds are fundamental to various applications, 
yet the extreme sparsity of high-precision geometric details hinders efficient context modeling, thereby limiting the compression speed and performance of existing methods.
To address this challenge, we propose a compact representation for efficient predictive lossless coding.
Our framework comprises two lightweight modules. First, the Geometry Re-Densification Module iteratively densifies encoded sparse geometry, extracts features at a dense scale, and then sparsifies the features for predictive coding. This module avoids costly computation on highly sparse details while maintaining a lightweight prediction head.
Second, the Cross-scale Feature Propagation Module leverages occupancy cues from multiple resolution levels to guide hierarchical feature propagation, enabling information sharing across scales and reducing redundant feature extraction.
Additionally, we introduce an integer-only inference pipeline to enable bit-exact cross-platform consistency, which avoids the entropy-coding collapse observed in existing neural compression methods and further accelerates coding.
Experiments demonstrate competitive compression performance at real-time speed.
Code is available at \href{https://github.com/pengpeng-yu/FastPCC}{https://github.com/pengpeng-yu/FastPCC}.
\end{abstract}

\renewcommand{\thefootnote}{\fnsymbol{footnote}}
\footnotetext[1]{Equal contribution.}
\footnotetext[2]{Corresponding author.}
\renewcommand{\thefootnote}{\arabic{footnote}}

\section{Introduction}
\label{sec:intro}

With the rapid advancement of 3D sensing technologies, massive amounts of point cloud data have been accumulated in various fields such as autonomous driving and mapping~\cite{you2020pseudo}. This surge in data volume has led to an increasing demand for precise point cloud compression (PCC). Currently, most PCC methods represent raw coordinate data using quantized structures such as range images~\cite{zhou2022riddle,wang2022point,stathoulopoulos2024recnet,li2026slide}, voxels~\cite{quach2019learning,he2022density,pang2022graspnet,wang2025Versatile,yu2025hierarchical}, or octrees~\cite{Biswas2020muscle,huang2020octsqueeze,que2021voxelcontext,chen2022point,fu2022octattention,song2023efficient,fu2026DeepRAHT}, and then apply techniques like prediction or transformation to achieve compression.

\begin{figure}[t]
    \begin{minipage}[b]{0.50\textwidth}
        \centering
        \subfloat[Level-8 octree of a KITTI point cloud]{\includegraphics[width=\linewidth,trim=-40 140 0 15,clip]{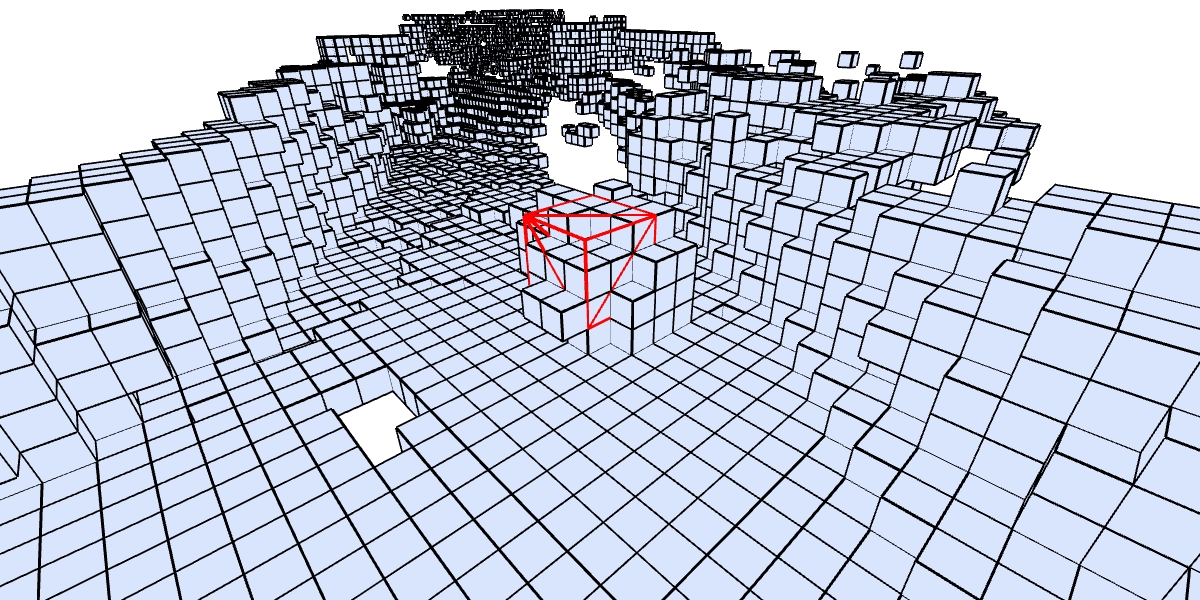}\label{fig_1a}}\\
        \vspace{0.25em}
        \subfloat[Level-12 octree of a KITTI point cloud]{\includegraphics[width=\linewidth,trim=-40 140 0 15,clip]{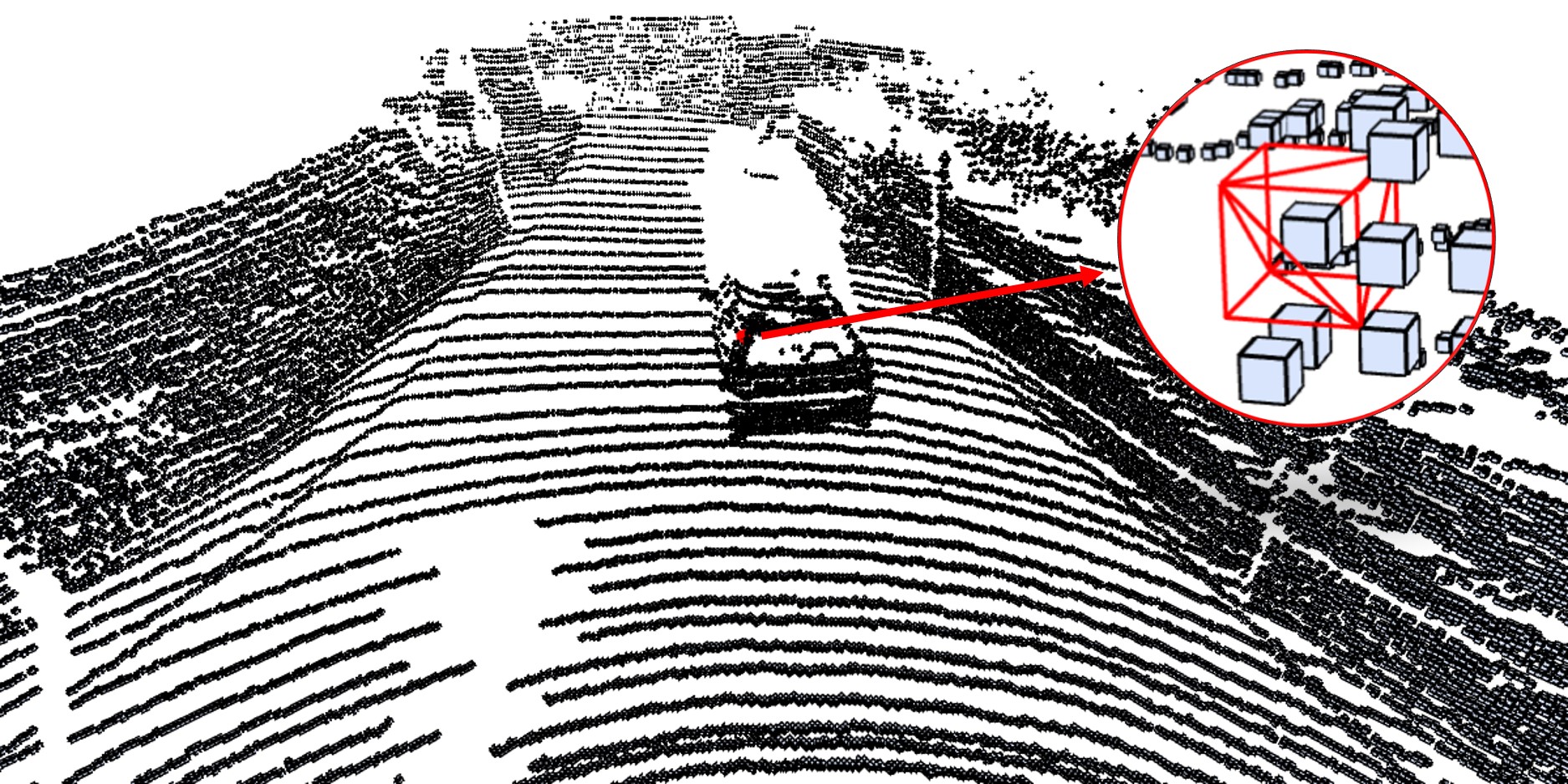}\label{fig_1b}}
    \end{minipage}
    \hfill
    \begin{minipage}[b]{0.44\textwidth}
        \centering
        \subfloat[Number of nodes and neighbors vs. level.]{\includegraphics[width=\linewidth,trim=6 9 -10 20,clip]{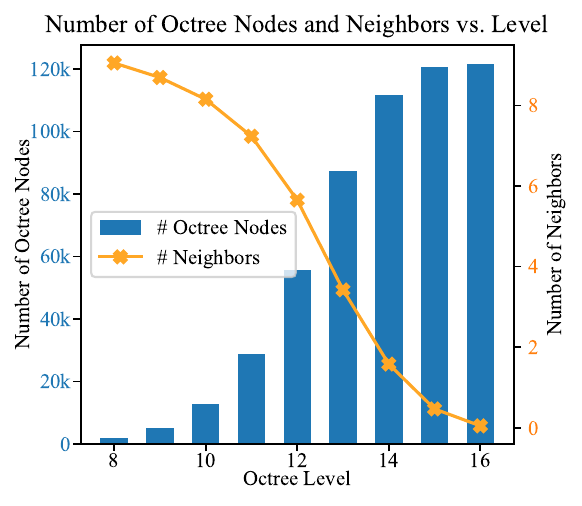}\label{Fig:HRCS}}
    \end{minipage}
    \caption{Illustration of the High-Resolution Contextual Sparsity (HRCS) phenomenon: (a) and (b) depict the voxelized octree representations at levels 8 and 12 for a point cloud from the KITTI dataset, respectively. The red bounding box highlights a $3 \times 3 \times 3$ neighborhood. (c) quantifies HRCS on the KITTI dataset, where the average number of neighbors per node decreases sharply with increasing octree level.}
    \vspace{-0.75em}
\end{figure}

Although existing PCC methods have made significant progress in rate-distortion~(RD) performance, their foundational representations, namely voxels or octrees, exhibit inherent limitations in high-precision compression scenarios. Both representations quantize a 3D space into discrete volumes, marking each as occupied only if it contains at least one point. However, as shown in Fig.~\ref{fig_1a} and Fig.~\ref{fig_1b}, as the quantization resolution increases, the local neighborhood around a given voxel becomes increasingly sparse, drastically reducing the availability of contextual information. We term this phenomenon \textbf{High-Resolution Contextual Sparsity (HRCS)}. In such cases, occupancy prediction becomes increasingly difficult due to the lack of local context.

To quantify HRCS, we conducted a statistical analysis on all frames of the KITTI dataset. For the octree of each sample, we collected two key statistics: (i) the total number of nodes at each level, and (ii) the average number of occupied neighbors within a $3\times3\times3$ neighborhood. As illustrated in Fig.~\ref{Fig:HRCS}, with increasing resolution (\ie, at deeper octree levels), 
the average number of neighbors per node drops sharply. At certain levels, the average number of neighbors even falls below one. 
Notably, this decline exhibits a marked inflection point at a specific octree level, indicating a nonlinear loss of contextual richness.

To address HRCS without sacrificing coding efficiency, we propose a Geometry Re-Densification (GRED) strategy. Instead of directly predicting sparse high-resolution nodes, GRED traces back to a shallower level, where geometry remains relatively dense, and performs dense-to-sparse feature transformation through lightweight convolutions and upsampling. The resulting features are spatially aligned with the target level and used to facilitate occupancy prediction. 
This process forms an iterative \emph{dense $\rightarrow$ sparse $\rightarrow$ predictive coding} cycle across octree levels, executed from coarse to fine resolutions.
Built upon GRED, we further introduce a Cross-Scale Feature Propagation (XFP) module to 
fuse dense features from shallow levels with sparse geometric cues from deeper levels, 
enabling effective cross-scale interaction while maintaining computational efficiency.


\begin{figure}[t]
  \centering
  \begin{minipage}[b]{0.5\textwidth}
      \centering
      \subfloat[BD-PSNR gains vs. the speed of encoding–decoding process on KITTI at 12-bit precision.]{
      \includegraphics[width=\linewidth,trim=0 -5 0 0,clip]{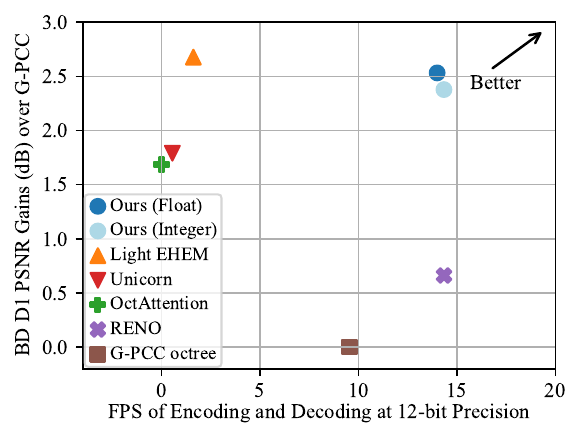}}
  \end{minipage}
  \hfil
  \begin{minipage}[b]{0.45\textwidth}
    \centering
    \subfloat[Non-deterministic floating-point inference leads to cross-platform decoding failure.]{
    \makebox[\linewidth][c]{
      \includegraphics[width=0.8\linewidth,trim=10 240 0 25,clip]{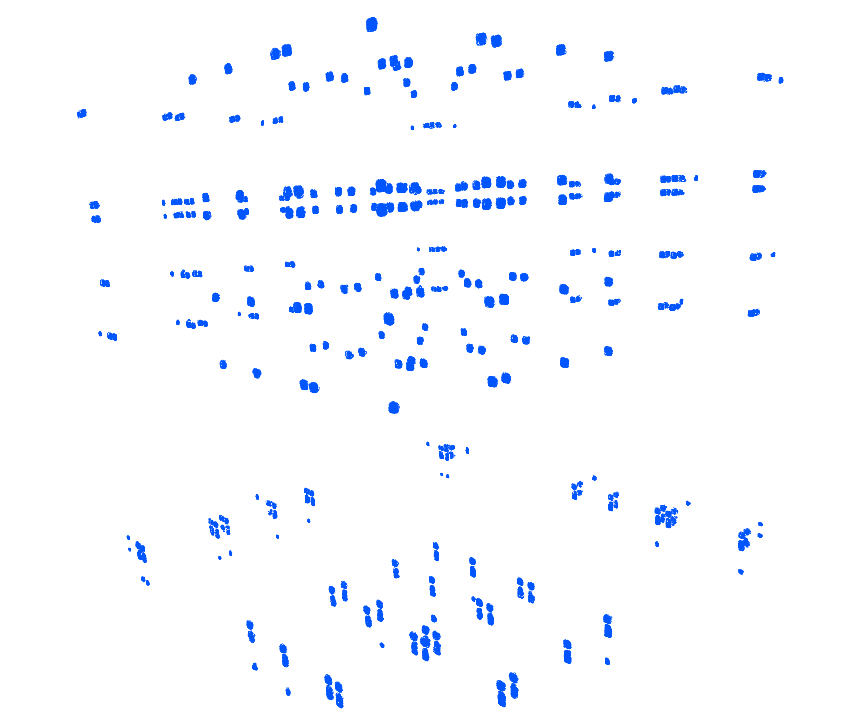}
    }}\\
    \vspace{0.5em}
    \subfloat[Integer-only inference enables bit-exact cross-platform decoding.]{
    \makebox[\linewidth][c]{
      \includegraphics[width=\linewidth,trim=180 130 0 215,clip]{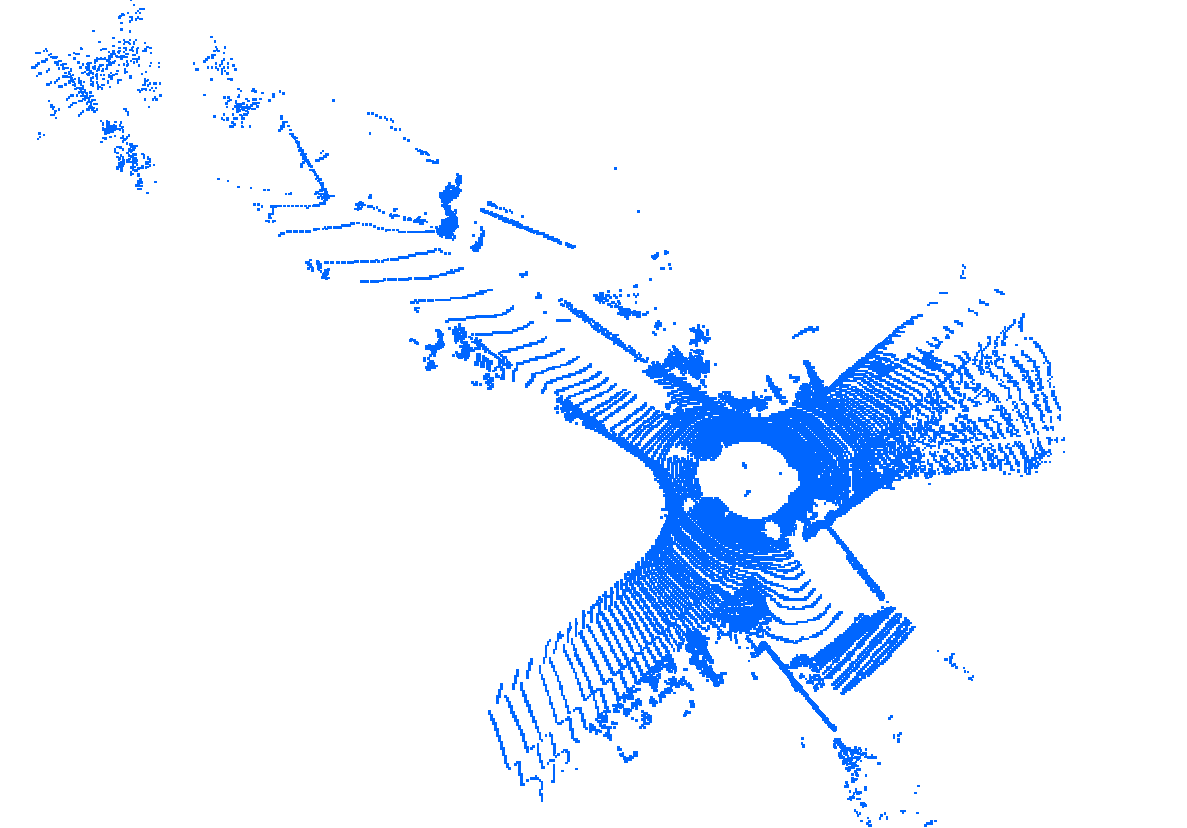}
    }
    }
  \end{minipage}
  \caption{
  (a) RD performance versus coding speed on KITTI at 12-bit precision, evaluated on an AMD EPYC 7R32 CPU and an NVIDIA RTX 4090 GPU.  
  (b) Decoding failure caused by non-deterministic floating-point computation, where encoding is performed on an NVIDIA RTX 4090 GPU and decoding on an RTX 5880 GPU. The reconstructed point cloud collapses into an approximately uniform distribution in 3D space.
  (c) Bit-exact decoding achieved with the proposed integer-only inference pipeline.
  }
  \vspace{-0.75em}
  \label{Fig:CodingTime_a}
\end{figure}

Beyond compression performance and coding latency, practical neural codecs must guarantee cross-platform bitstream decodability. However, floating-point inference is inherently non-deterministic across heterogeneous devices and software stacks, and even minor numerical discrepancies in probability prediction may catastrophically break entropy coding. To address this practical challenge, we further develop an integer-only inference pipeline, ensuring bit-exact cross-platform consistency while enabling efficient execution.

In the following sections, we first review related work in PCC, followed by a detailed description of GRED, XFP, and our integer-only inference pipeline. Extensive experiments on KITTI~\cite{geiger2012are} and Ford~\cite{pandey2011ford} demonstrate the effectiveness of the proposed method in terms of compression performance, coding efficiency, and cross-platform consistency.

\section{Related work}
This section reviews representative PCC methods and briefly discusses recent advances in cross-platform consistency for neural codecs. According to the underlying geometric representation, existing PCC approaches can be broadly categorized into two groups: (1) voxel-based PCC and (2) octree-based PCC.

\textbf{Voxel-based PCC}. Voxel-based approaches split the point cloud into sufficiently small voxels, utilizing sparse convolution~\cite{tang2023torchsparse} to optimize memory usage. Based on the voxel, many PCC techniques have emerged~\cite{wiesmann2021deep,nguyen2021lossless,tzamarias2022fast,nguyen2022learning,pang2024pivot,zhang2025scalable,meng2025pcgcd,zhang2025rate}. 
For example, Wang~\etal~\cite{wang2021lossy} proposed a voxel-based geometry compression method that partitions point clouds into non-overlapping 3D cubes and leverages a variational autoencoder-driven convolutional neural network to extract latent features and hyperpriors for entropy coding. 
Recently, Wang~\etal~\cite{wang2025Versatile} proposed a universal multiscale conditional coding framework, Unicorn, which leverages sparse tensors from voxelized point clouds and cross-scale temporal priors to enhance geometry compression. 
Zhang~\etal~\cite{zhang2025adadpcc} proposed a dynamic point cloud compression framework based on voxelized data, featuring a slimmable architecture with multiple coding routes for rate-distortion optimization, and a coarse-to-fine motion module to improve inter-frame prediction.

\textbf{Octree-based PCC}. Octree-based approaches typically construct an $L$ level octree by recursively subdividing the point cloud within a pre-defined bounding volume, and achieve compression by predicting the occupancy status of each octree node. Based on the octree structure, many PCC techniques have emerged~\cite{kammerl2012real,golla2015real,garcia2017context,wen2020lossy,luo2024scp,wang2025topnet}. For example, Huang~\etal~\cite{huang2020octsqueeze} proposed an octree-based compression method that leverages a tree-structured conditional entropy model to exploit sparsity and structural redundancy in LiDAR point clouds. Similarly, Fu~\etal~\cite{fu2022octattention} proposed an octree-based deep learning framework that encodes point clouds by leveraging rich sibling and ancestor contexts with an attention mechanism. Cui~\etal~\cite{cui2023octformer} proposed OctFormer, which constructs node sequences with non-overlapping context windows and shares attention results to reduce computation. Song~\etal~\cite{song2023efficient} proposed an octree-based entropy model with a hierarchical attention mechanism and grouped context structure, reducing the complexity and decoding latency of large-scale auto-regressive models.

\textbf{Cross-platform consistency.}
Prior studies in neural image/video compression reveal that even tiny floating-point computation errors can cause encoder-decoder mismatch and lead to decoding failures~\cite{Balle2019IntegerNetworks,Koyuncu2022DeviceInteroperability,Tian2023CalibrationNVC,Conceicao2025CrossPlatformCaseStudy,Jia2025DCVCRT}.
This issue typically stems from entropy coding, which requires the encoder and decoder to share exactly the same probability distribution.
A principled solution is model integerization, which enforces bit-exact deterministic computations~\cite{Balle2019IntegerNetworks,Koyuncu2022DeviceInteroperability,Conceicao2025CrossPlatformCaseStudy,Jia2025DCVCRT}.
Alternative approaches transmit auxiliary calibration information to correct cross-platform mismatch, at the cost of extra side information~\cite{Tian2023CalibrationNVC}.
Compared with image/video codecs, neural PCC often involves sparse operators and data-dependent control flow, which complicates deterministic deployment across heterogeneous platforms.
While cross-platform consistency is increasingly studied for image/video neural codecs, it remains largely underexplored in neural PCC.

\textbf{In summary}, both voxel-based and octree-based PCC approaches have achieved remarkable progress, with learning-based methods significantly improving RD performance over traditional codecs~\cite{mekuria2017three,schwarz2019emerging, garcia2020geometry,song2021layer,wang2022rpcc,qin2024multi,cao2025real}. 
However, under high-resolution settings, both representations tend to suffer from HRCS. 
Moreover, existing PCC studies primarily focus on RD performance, while the practical requirement of cross-platform deterministic coding remains largely overlooked. This motivates our work to jointly address HRCS and cross-platform consistency within a unified neural PCC framework.

\section{Method}

\begin{figure}[t]
  \centering
  \includegraphics[width=\linewidth,trim=45 22 45 70,clip]{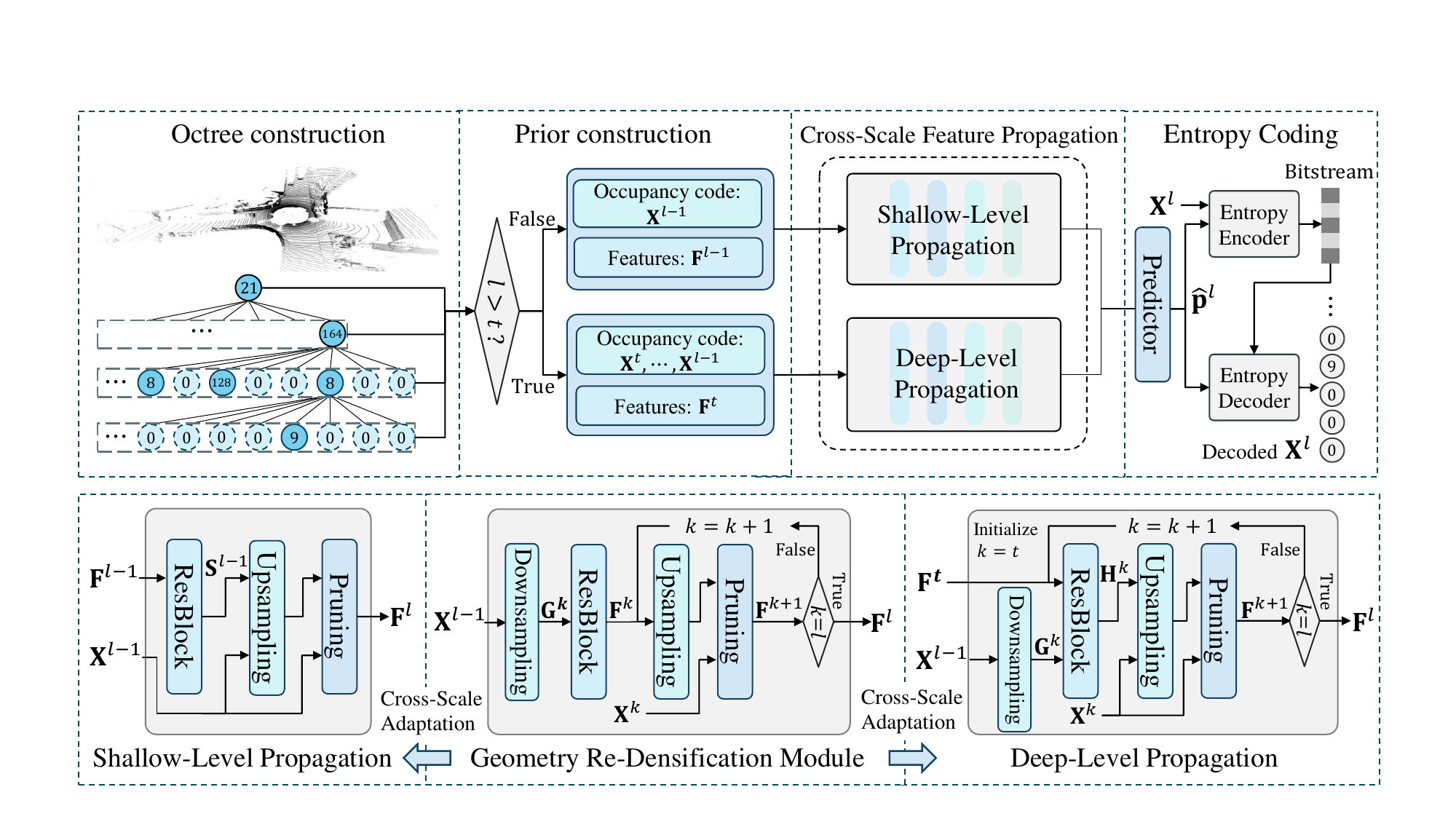}
  \caption{Pipeline of compressing a single octree level in the proposed lossless LiDAR PCC framework. The framework consists of four main stages: octree construction, prior construction, cross-scale feature propagation, and entropy coding. The cross-scale feature propagation module comprises two key components: the shallow-level propagation block and the deep-level propagation block, both adapted from the geometry re-densification module to exploit cross-scale features.}
  \vspace{-1em}
  \label{fig_2}
\end{figure}

To address the HRCS problem while satisfying the practical requirements of real-time LiDAR PCC, we propose a fast octree-based encoding framework. The overall architecture is illustrated in Fig.~\ref{fig_2}. The proposed framework consists of four key components: octree construction, prior construction, cross-scale feature propagation, and entropy coding. In the following, we first introduce the Cross-Scale Feature Propagation module with its core Geometry Re-Densification design, and then describe the integer-only inference pipeline that guarantees cross-platform consistent coding.

\subsection{Geometry Re-Densification Module}
The irregular and unordered nature of LiDAR point clouds poses significant challenges for efficient processing on modern hardware architectures.
To better exploit existing hardware, most compression methods convert raw point clouds into octree structures. By recursively dividing space into eight subcells at each level, octrees provide a compact and hierarchical representation of geometry. The maximum level $L$ of the octree controls reconstruction fidelity.
Using octrees, existing codecs can perform progressive, dense-to-sparse predictive lossless coding of occupancy codes, modeling the distribution by exploiting contexts from encoded sibling nodes and ancestral nodes to minimize storage.


Despite the compact and regular structure of octree representations, they face the challenge of HRCS in encoding high-resolution LiDAR point clouds. 
To address this problem, we propose a \textbf{Geometry Re-Densification (GRED)} module and integrate it into the dense-to-sparse progressive coding pipeline. 
At each HRCS-affected level, GRED downsamples sparse occupancy codes to obtain denser contextual features, and then reverts to the original sparse domain for predictive coding.
For each HRCS-affected level, the module performs:

\begin{enumerate}
  \item \emph{Re-Densification.} Downsample the occupancy codes of the last encoded level into a denser octree level, producing an aligned dense feature map with zero-padding for empty nodes.

  \item \emph{Feature Extraction.} Apply lightweight convolutions to the dense feature map to extract rich local spatial representations.

  \item \emph{Re-Sparsification.} Recursively upsample and prune the dense features using the encoded occupancy codes, producing a sparse feature map aligned with the nodes at the current level.

  \item \emph{Prediction \& Coding.} Use a multilayer perceptron-based predictor on the sparse feature map to estimate the occupancy distribution over 255 classes (8-bit occupancy patterns), and entropy-encode the occupancy codes.
\end{enumerate}

Without loss of generality, we denote octree as $\mathbf{X} = \{\mathbf{X}^1, \mathbf{X}^2, \dots, \mathbf{X}^L\}$, where $L$ is the maximum level of octree, $\mathbf{X}^l \in \{1, \dots, 255\}^{N^l}$ represents the occupancy sequence of all nodes at level $l$, and $N^l$ denotes the number of nodes at that level.
Lossless compression approximates the true occupancy distribution $P(\mathbf{X})$ with an estimated distribution $Q(\mathbf{X})$ by minimizing the cross-entropy:
\begin{equation}
  H(P,Q) = \mathbb{E}_{P(\mathbf{X})}\left[-\log Q(\mathbf{X})\right].
\end{equation}

Standard octree-based codecs typically estimate the distribution of occupancy codes in a layer-wise autoregressive manner, \ie, previously encoded levels serve as priors for predicting the current one:
\begin{equation}
  Q(\mathbf{X}) = \prod_{l=1}^L Q\left(\mathbf{X}^l \mid \mathbf{X}^{1:l-1}\right),
\end{equation}
where each conditional distribution is predicted by an occupancy predictor:
\begin{equation}
  Q\left(\mathbf{X}^l \mid \mathbf{X}^{1:l-1}\right) = \mathrm{Predictor}\left(\mathbf{X}^{1:l-1}\right).
\end{equation}


Suppose predictions are being made at level $l$. 
Given the encoded occupancy codes $\{\mathbf{X}^1,\cdots,\mathbf{X}^{l-1}\}$,
to obtain denser context features, GRED first downsamples $\mathbf{X}^{l-1}$ into a pre-defined dense octree level $k$:
\begin{equation}
  \mathbf{G}^k = \mathrm{Downsampling}\left(\mathbf{X}^{l-1}\right),
\end{equation}
where the $\mathrm{Downsampling}(\cdot)$ operation embeds the occupancy codes of $l-1$ into feature maps at level $k$ using sparse convolutions. In $\mathbf{G}^k$, each channel corresponds to a specific occupancy state of a node at level $l-1$, thereby enriching contextual information with higher density.

To extract contextual features, $\mathbf{G}^k$ is fed to a $\mathrm{ResBlock}$ for \emph{Feature Extraction}, thereby obtaining the feature $\mathbf{F}^k$:
\begin{equation}
  \mathbf{F}^k = \mathrm{ResBlock}\left(\mathbf{G}^k\right).
\end{equation}

Although it is possible to directly predict occupancy in this dense space, it would incur prohibitive computational costs due to the vast number of potential sub-nodes. Instead, GRED progressively reverts $\mathbf{F}^k$ to the original sparse space $\mathbf{F}^{l}$ through multi-step upsampling, thereby achieving the \emph{Re-Sparsification} of the features:
\begin{equation}
  \mathbf{F}^{k+1} = \mathrm{Pruning}\left(\mathrm{Upsampling}\left(\mathbf{F}^k\right),\mathbf{X}^{k}\right),
\end{equation}
where $\mathrm{Upsampling}(\cdot)$ is a linear transformation followed by a PReLU activation, performing an 8$\times$ channel expansion, and $\mathrm{Pruning}(\cdot)$ discards features of unoccupied child nodes.
This process is recursively applied until the feature map $\mathbf{F}^{l}$ is obtained.
The feature is then fed into an MLP-based predictor to estimate the occupancy distribution:
\begin{equation}
  \hat{\mathbf{p}}^l = \mathrm{Predictor}\left(\mathbf{F}^{l}\right).
\end{equation}

Then the true occupancy codes $\mathbf{X}^l$ are entropy-encoded using $\hat{\mathbf{p}}^l$, finishing \emph{Prediction \& Coding}.
Overall, GRED enriches contextual information with low computational overhead while preserving the progressive decoding workflow.

Although many 3D tasks employ densification operations~\cite{choe2022deep,deng2024linnet}, such as quantization and downsampling, before processing and analysis, LiDAR point cloud compression presents a unique constraint: the decoder cannot access the full geometry at the beginning of decoding. Therefore, globally pre-densifying all octree levels is infeasible. This insight, combined with the observed nonlinear drop in occupancy density across octree levels, supports the necessity of on-the-fly re-densification within a progressive octree coding pipeline.

\subsection{Cross-Scale Feature Propagation Module}
While the proposed GRED module effectively mitigates HRCS, we delve into the rich inter‐scale contextual dependencies across the octree.
Existing octree‐based codecs typically extract features and predict occupancy codes independently at each octree level or within local node windows. However, this per‐level processing overlooks the strong contextual dependencies across octree levels, leading to redundant feature extraction.

To fully leverage inter‐scale context, we propose a unified \textbf{Cross‐Scale Feature Propagation (XFP) Module} that (i) directly propagates features across octree levels and (ii) generalizes the core idea of the GRED Module into a broader, multi-scale framework. In fact, the GRED Module can be viewed as a special case of XFP, applied only at the deepest levels. XFP shares features from coarser (shallower) levels with finer (deeper) levels, avoiding redundant feature extraction and enhancing contextual awareness.


Suppose we are predicting the occupancy codes at level $l$, meaning that the preceding feature maps $\{\mathbf{F}^1,\cdots,\mathbf{F}^{l-1}\}$ and occupancy codes $\{\mathbf{X}^1,\cdots,\mathbf{X}^{l-1}\}$ are available. The first step in XFP is to determine an appropriate feature propagation strategy. In this work, we define two propagation regimes based on a pre-defined threshold level $t$:
\begin{enumerate}
  \item \textbf{Shallow levels} ($l \leq t$): feature propagation is conducted without \ re-densification, as the geometry remains relatively dense.
  \item \textbf{Deep levels} ($l > t$): feature propagation incorporates contextual information from the dense level $k$ through occupancy-based re-densification.
\end{enumerate}

\textbf{Shallow-Level Propagation}.
For levels $l \le t$, the octree is relatively shallow and the contextual information is sufficiently dense, making the HRCS problem less prominent. 
At these levels, we adopt a simplified version of the GRED module, omitting the \emph{Re-Densification} step. 

In the \emph{Feature Extraction} step, since the re-densified feature $\mathbf{G}$ is omitted, the input is directly the encoded feature from level $l{-}1$, denoted as $\mathbf{F}^{l-1}$. A $\mathrm{ResBlock}$ is then applied to extract features, obtain the representation $\mathbf{S}^{l-1}$:
\begin{equation}
  \mathbf{S}^{l-1} = \mathrm{ResBlock}(\mathbf{F}^{l-1}).
\end{equation}

Next, through one step of \emph{Re-Sparsification}, $\mathbf{S}^{l-1}$ is upsampled to level $l$ to produce the re-sparsified feature $\mathbf{F}^l$:
\begingroup
\setlength{\thinmuskip}{0.5mu}
\setlength{\medmuskip}{1.mu}
\setlength{\thickmuskip}{1.mu}
\begin{equation}
  \mathbf{F}^{l} = \mathrm{Pruning}\left(\mathrm{Upsampling}\left(\mathrm{Concat}\left(\mathbf{S}^{l-1}, \mathbf{X}^{l-1}\right)\right),\;\mathbf{X}^{l-1}\right),
\end{equation}
\endgroup
where $\mathrm{Concat}$ denotes channel-wise concatenation of matrices. The obtained feature $\mathbf{F}^l$ then undergoes the same \emph{Prediction \& Coding} as GRED. 

\textbf{Deep-Level Propagation with Re-Densification}.
For $l > t$, the spatial sparsity makes direct propagation less effective. Therefore, we apply the full GRED module at these levels. During this process, we incorporate additional inter-scale contextual information to enrich the extracted features. 

GRED utilizes the re-densified feature $\mathbf{G}^k$ for \emph{Feature Extraction}. To fully leverage the information from the previous scale, we concatenate $\mathbf{G}^k$ with the original feature map $\mathbf{F}^k$ and use $\mathrm{ResBlock}$ to obtain the fused representation: 
\begin{equation}
  \mathbf{H}^k = \mathrm{ResBlock}\left(\mathrm{Concat}(\mathbf{F}^k,\;\mathbf{G}^k)\right).
\end{equation}

The fused representation $\mathbf{H}^k$ replaces the original input feature $\mathbf{F}^k$ in the \emph{Re-Sparsification}, 
enabling multi-scale features to be fused into the current level:
\begingroup
\setlength{\thinmuskip}{1mu}
\setlength{\medmuskip}{2mu}
\setlength{\thickmuskip}{2mu}
\begin{align}
  \mathbf{F}^{k+1} &=  \mathrm{Pruning}\left(\mathrm{Upsampling}\left(\mathrm{Concat}(\mathbf{H}^k,\mathbf{X}^k)\right),\;\mathbf{X}^{k}\right), 
  \quad k = t, \dots, l-1. 
\end{align}
\endgroup

By recursively applying the above process, we obtain the fused feature $\mathbf{F}^l$.
Finally, \emph{Prediction \& Coding} is performed at level $l$ based on the feature $\mathbf{F}^l$.

This cross-scale propagation scheme reuses context-rich features from earlier levels and adapts them to finer resolutions through sparse, occupancy-aware operations. 
By combining shallow and deep propagation pathways, XFP unifies dense and sparse processing, enabling efficient feature extraction throughout the octree hierarchy.

\subsection{Integer-Only Inference for Cross-Platform Consistency}

Neural compression models require strict numerical consistency between the encoder and the decoder, since entropy coding is highly sensitive to numerical errors. Even minor discrepancies in predicted probabilities may lead to decoding failures~\cite{Balle2019IntegerNetworks,Jia2025DCVCRT,Tian2023CalibrationNVC,Conceicao2025CrossPlatformCaseStudy}.
However, floating-point computations are inherently non-deterministic across hardware and software. 
In practical deployment, enforcing identical execution environments is unrealistic. 
To enable reliable cross-platform coding, we therefore develop an integer-only inference pipeline that guarantees deterministic computation.
To the best of our knowledge, this is the first cross-platform integer-only inference pipeline for neural PCC.

Our objective is to eliminate all floating-point operations during inference while preserving both runtime efficiency and compression performance.
To this end, we adopt a mixed-precision integer design that balances computational cost and numerical stability:
(i) computation-intensive operators are executed using low-bit integer arithmetic for efficiency;
(ii) lightweight yet numerically sensitive operators are implemented with higher-precision fixed-point arithmetic to control quantization error.
This design enables fully integerized end-to-end deterministic execution without sacrificing much performance.

\paragraph{Quantization.}
We employ 8-bit integer quantization for compute-dominant operators, including linear layers and sparse convolutions, while using 32-bit fixed-point arithmetic for re-quantization and activation functions.
Given a floating-point tensor $\mathbf{x}$, its quantized integer representation is defined as
\begin{equation}
\mathbf{q}_x = \mathcal{Q}(\mathbf{x}; s, z)
= \mathrm{clip}\!\left(
\left\lfloor \frac{\mathbf{x}}{s} \right\rceil + z,\;
q_{\min}, q_{\max}
\right),
\end{equation}
where $s$ and $z$ denote the scale and zero-point, respectively, and $\lfloor \cdot \rceil$ represents a rounding operation.
Quantization parameters for activations in linear and sparse convolution operators are obtained via lightweight calibration on a small subset of training data.
All remaining operators are implemented in fixed-point format and therefore do not require distribution statistics.
During inference, all computations are performed directly in integer space, ensuring deterministic behavior.

\paragraph{Integer linear and sparse convolution.}
For linear layers, activations and weights are represented as 8-bit integers, while accumulation operations are performed using 32-bit integers:
\begin{equation}
\mathbf{y}_{\mathrm{int32}}
=
\sum_{c}
(\mathbf{q}_x - z_x)\,(\mathbf{q}_w - z_w)^\mathrm{T}
+ \mathbf{b}_{\mathrm{int32}},
\end{equation}
where $\mathbf{q}_x$ and $\mathbf{q}_w$ denote quantized activations and weights, $z_x$ and $z_w$ are the corresponding zero-points, and $\mathbf{b}_{\mathrm{int32}}$ denotes the quantized bias.
In our implementation, weights are symmetrically quantized (\ie, $z_w = 0$).
The accumulated result is then re-quantized to 8-bit representation via fixed-point scaling:
\begin{equation}
\mathbf{q}_y =
\mathrm{clip}\!\left(
\left\lfloor
\mathbf{y}_{\mathrm{int32}} \cdot m / 2^{r}
\right\rceil
+ z_y, \;
q_{\min}, q_{\max}
\right),
\end{equation}
where $m$ is a precomputed integer multiplier, $r$ is a fixed right-shift factor, and $z_y$ denotes the output zero-point.

For sparse convolution, each operator is decomposed into multiple indexed linear transforms~\cite{choy20194d}.
This decomposition preserves sparsity while enabling efficient execution using integer linear primitives.

\paragraph{Integer softmax.}
The occupancy predictor outputs a 255-way probability distribution via a softmax operation that is directly consumed by entropy coding.
To avoid platform-dependent floating-point exponentiation and division, we implement softmax entirely in fixed-point arithmetic.
Given the logits $\boldsymbol{\ell}$, the softmax probability is computed in a numerically stable form as
\begin{equation}    
p_i =
\frac{\exp(\ell_i - \max_k \ell_k)}
{\sum_{j} \exp(\ell_j - \max_k \ell_k)}, \,\,\,\, i,j,k \in \{1,\dots,255\}.
\end{equation}

The exponential function is approximated using a precomputed lookup table that only covers the non-positive domain (i.e., $\ell_i - \max_k \ell_k \le 0$), which significantly reduces table size while preserving numerical stability.
Accumulation and normalization are performed using 32-bit integer arithmetic.
The resulting probabilities are represented in high-precision fixed-point format, guaranteeing bit-exact consistency across different platforms.

\section{Experiments}
In this section, we present a comprehensive experimental evaluation of our method, including experimental settings, comparative results with state-of-the-art approaches, and ablation studies.

\subsection{Settings}
In this section, we detail the experimental setup, including the benchmark datasets, evaluation metrics, and comparative baselines.

\textbf{Benchmark Datasets}. Experiments are conducted on two different LiDAR datasets: KITTI~\cite{geiger2012are} and Ford~\cite{pandey2011ford}. The KITTI dataset consists of 22 stereo sequences collected by a Velodyne LiDAR scanner across diverse continuous scenes, totaling 43,552 frames. Following Fu~\etal~\cite{fu2022octattention}, we use sequences $\#00$ to $\#10$ for training and $\#11$ to $\#21$ for testing. The Ford dataset comprises three sequences, each containing 1,500 frames. Consistent with Song~\etal~\cite{song2023efficient}, we use sequence $\#01$ for training, and sequences $\#02$ and $\#03$ for testing. 

\textbf{Evaluation Metrics}. We adopt point-to-point PSNR (D1 PSNR) and point-to-plane PSNR (D2 PSNR)~\cite{tian2017geometric} as distortion measures. These are standard metrics recommended by MPEG~\cite{schwarz2019emerging}. 
We employ the Bj{\o}ntegaard Delta (BD) metrics~\cite{bjontegaard2001calculation} for evaluating rate-distortion performance, namely Bj{\o}ntegaard Delta Peak Signal-to-Noise Ratio (BD-PSNR) and Bj{\o}ntegaard Delta Rate (BD-Rate).
It is important to note that both BD-Rate and BD-PSNR measure the \textbf{relative gains} of a tested model compared to a baseline.
A negative BD-Rate or a positive BD-PSNR indicates that the tested model outperforms the baseline.


\textbf{Compared Methods}. 
We compare the proposed framework with 5 representative PCC methods. 
G-PCC~\cite{schwarz2019emerging}, standardized by MPEG, serves as the classical geometry-based benchmark. 
OctAttention~\cite{fu2022octattention} and Light EHEM~\cite{song2023efficient} are transformer-based octree compression methods, while Unicorn~\cite{wang2025Versatile} represents a recent voxel-based PCC framework. 
In addition, RENO~\cite{you2025reno} introduces an efficient sampling strategy to balance compression performance and computational efficiency.
Unless otherwise specified, all methods are re-evaluated on a computer equipped with an AMD EPYC 7R32 CPU and an NVIDIA RTX 4090 GPU.
Due to the unavailability of the official implementation of Light EHEM, we report the performance metrics as provided in the original paper.

\subsection{Performance Analysis}
This section evaluates the proposed method in terms of rate-distortion performance and computational efficiency, providing a comprehensive assessment of its effectiveness. 

\begin{figure}[tb]
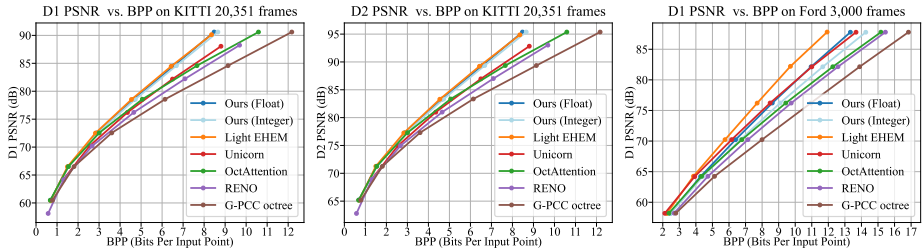

  \centering
  \includegraphics[width=0.328\linewidth,trim=5 6 5 5,clip]{data/KITTI/sample-wise D1 PSNR (dB)/D1 PSNR (dB) KITTI.pdf}
  \includegraphics[width=0.328\linewidth,trim=5 6 5 5,clip]{data/KITTI/sample-wise D2 PSNR (dB)/D2 PSNR (dB) KITTI.pdf}
  \includegraphics[width=0.328\linewidth,trim=5 6 5 5,clip]{data/Ford/sample-wise D1 PSNR (dB)/D1 PSNR (dB) Ford.pdf}\label{Fig:RD_c}
  \caption{RD performance comparison with existing methods on KITTI and Ford datasets from 11 bits to 16 bits.}  
  \label{Fig:RD}
\end{figure}

\begin{table}[tb]
  \centering
  \caption{BD-Rate (\%) and BD-PSNR (dB) gains of our floating-point model.}
  \setlength{\tabcolsep}{2.6pt}
  \fontsize{8}{10}\selectfont
  \begin{tabular}{lrrrrrrrr}
\toprule
Ours (Float) vs. & \multicolumn{4}{c}{KITTI} & \multicolumn{4}{c}{Ford} \\
Existing & \multicolumn{2}{c}{BD-Rate} & \multicolumn{2}{c}{BD-PSNR} & \multicolumn{2}{c}{BD-Rate} & \multicolumn{2}{c}{BD-PSNR} \\
Methods & D1  & D2  & D1  & D2  & D1  & D2  & D1  & D2 \\
\midrule
OctAttention~\cite{fu2022octattention} (AAAI'22) & -7.294 & -7.332 & 0.754 & 0.760 & -6.205 & -6.193 & 0.969 & 0.969 \\
Light EHEM~\cite{song2023efficient} (CVPR'23) & 1.429 & 1.422 & -0.152 & -0.152 & 11.535 & 11.987 & -1.899 & -1.960 \\
Unicorn~\cite{wang2025Versatile} (TPAMI'25) & -10.862 & -10.889 & 1.367 & 1.373 & 3.922 & 3.989 & -0.622 & -0.637 \\
RENO~\cite{you2025reno} (CVPR'25) & -15.582 & -15.579 & 1.887 & 1.890 & -11.049 & -11.040 & 1.938 & 1.938 \\
G-PCC octree~\cite{schwarz2019emerging} & -21.931 & -21.954 & 2.532 & 2.541 & -19.150 & -19.143 & 3.411 & 3.412 \\
\bottomrule
\end{tabular}%
  \label{Tab:BDGain}%
\end{table}%

\begin{table}[tbp]
  \centering
  \caption{BD-Rate (\%) and BD-PSNR (dB) gains of our integer-only model.}
  \setlength{\tabcolsep}{4.2pt}
  \fontsize{8}{10}\selectfont
\begin{tabular}{lrrrrrrrr}
\toprule
Ours (Integer) vs. & \multicolumn{4}{c}{KITTI} & \multicolumn{4}{c}{Ford} \\
Existing Real-time & \multicolumn{2}{c}{BD-Rate} & \multicolumn{2}{c}{BD-PSNR} & \multicolumn{2}{c}{BD-Rate} & \multicolumn{2}{c}{BD-PSNR} \\
Methods & D1  & D2  & D1  & D2  & D1  & D2  & D1  & D2 \\
\midrule
RENO~\cite{you2025reno} (CVPR'25) & -14.289 & -14.285 & 1.728 & 1.730 & -6.875 & -6.864 & 1.185 & 1.185 \\
G-PCC octree~\cite{schwarz2019emerging} & -20.705 & -20.728 & 2.378 & 2.387 & -15.355 & -15.348 & 2.694 & 2.694 \\
\bottomrule
\end{tabular}%
  \label{Tab:BDGainIntegerOnly}%
\end{table}%

\textbf{Rate-Distortion Performance}. 
This section presents the RD performance of the proposed method compared to several existing methods, using two standard evaluation curves: D1 PSNR vs. Bits Per input Point (BPP) and D2 PSNR vs. BPP. 
A curve closer to the upper-left corner indicates higher geometry precision at lower bitrates, reflecting better compression performance. 

The experimental results are illustrated in Fig.~\ref{Fig:RD}. On the KITTI dataset, our method achieves performance comparable to the transformer-based Light EHEM and outperforms the sparse convolution-based Unicorn, demonstrating clear advantages in RD performance while providing substantially improved computational efficiency. Notably, the integer-only model maintains RD performance comparable to its floating-point counterpart. On the Ford dataset, the overall RD performance is relatively less competitive, yet our method still surpasses approaches with similar coding latency, such as RENO and G-PCC octree. This performance gap is likely due to the limited number of training samples (1,500 frames), which constrains generalization capability. Table~\ref{Tab:BDGain} presents the quantitative BD-rate and BD-PSNR metrics of the proposed method compared with existing approaches, and Table~\ref{Tab:BDGainIntegerOnly} reports the corresponding results of the integer-only model over real-time baselines. On the KITTI dataset, compared with the efficiency-oriented method RENO, our approach achieves gains of 1.887 dB and 1.890 dB in D1 and D2 PSNR, respectively, while the integer-only model obtains gains of 1.728 dB and 1.730 dB under the same setting. These results demonstrate that the proposed framework effectively exploits structural redundancy within the octree representation, achieving competitive compression performance together with high runtime efficiency and strict cross-platform consistency.

{
\begin{table}[tb]
  \centering
  \caption{
    Comparison of coding time across 11--16 bits with existing real-time methods.}
    \fontsize{8}{10}\selectfont
    \setlength\tabcolsep{5pt}
\begin{tabular}{llrrrr}
\toprule
\multirow{2}[2]{*}{Methods} & \multicolumn{1}{c}{\multirow{2}[2]{*}{Device}} & \multicolumn{2}{c}{KITTI} & \multicolumn{2}{c}{Ford}  \\
    &     & \multicolumn{1}{c}{Enc Time} & \multicolumn{1}{c}{Dec Time} & \multicolumn{1}{c}{Enc Time} & \multicolumn{1}{c}{Dec Time} \\
\midrule
G-PCC octree & AMD EPYC 7R32 & 0.149s & 0.103s & 0.150s & 0.107s \\
RENO & NVIDIA RTX 4090 & 0.059s & 0.056s & 0.072s & 0.057s \\
Ours (Float) & NVIDIA RTX 4090 & 0.075s & 0.081s & 0.093s & 0.103s \\
Ours (Integer) & NVIDIA RTX 4090 & 0.060s & 0.070s & 0.067s & 0.082s \\
Ours (Integer) & NVIDIA RTX 5880 & 0.047s & 0.056s & 0.054s & 0.067s \\
Ours (Integer) & NVIDIA Tesla V100 & 0.166s & 0.175s & 0.187s & 0.199s \\
\bottomrule
\end{tabular}%
\label{Tab:CodingTime}%
\end{table}%
}

\begin{figure}
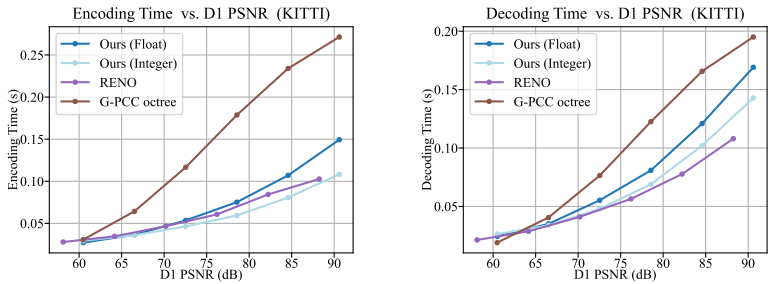

  \centering
  \includegraphics[width=0.42\linewidth,trim=0 6 -15 0,clip]{data/KITTI/sample-wise Encoding Time (s)/Encoding Time (s) KITTI.pdf}
  \includegraphics[width=0.42\linewidth,trim=-15 6 0 0,clip]{data/KITTI/sample-wise Decoding Time (s)/Decoding Time (s) KITTI.pdf}
  \vspace{-0.5em}
  \caption{Comparison of coding time across 11--16 bits with existing real-time methods.}
  \label{Fig:CodingTime}
\end{figure}

\textbf{Computational Efficiency}. 
To evaluate the real-time capability of the proposed framework, we measure the encoding and decoding time of our method and several representative methods, as summarized in Table~\ref{Tab:CodingTime}. 
The proposed method demonstrates faster runtime than most competing methods. For a clearer comparison, Fig.~\ref{Fig:CodingTime_a}(a) illustrates the BD-PSNR gains versus frames per second (FPS) on the 12-bit quantized KITTI dataset. Our method achieves 14 FPS for the complete encoding–decoding process, surpassing other methods with similar compression performance. 
To further analyze runtime behavior under varying geometry precisions, we compare the averaged encoding and decoding time across 11--16-bit quantization settings against existing real-time methods. As shown in Fig.~\ref{Fig:CodingTime}, the integer-only model consistently outperforms G-PCC, while maintaining runtime comparable to the efficiency-oriented RENO. These results confirm that the proposed framework achieves a favorable balance between compression performance and computational efficiency. 

In addition, Table~\ref{Tab:CodingTime} reports the averaged runtime of our integer-only model across multiple GPUs beyond the default RTX 4090, including RTX 5880 and Tesla V100. The integer-only pipeline achieves robust throughput on different devices, while its runtime varies with hardware capability as expected (\eg, faster on RTX 5880 and slower on V100). These results further validate the practicality of our cross-platform design for deployment on heterogeneous platforms.


\begin{figure}[tb]
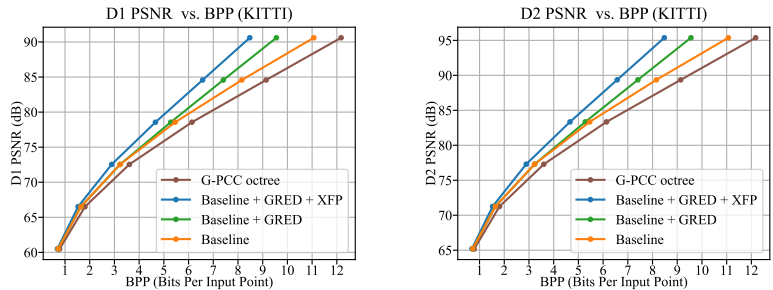

  \centering
  \includegraphics[width=0.42\linewidth,trim=0 6 -15 0,clip]{data/Ablation/sample-wise D1 PSNR (dB)/D1 PSNR (dB) KITTI.pdf}
  \includegraphics[width=0.42\linewidth,trim=-15 6 0 0,clip]{data/Ablation/sample-wise D2 PSNR (dB)/D2 PSNR (dB) KITTI.pdf}
  \vspace{-0.5em}
  \caption{RD performance comparison of ablated models from 11 bits to 16 bits.}
  \vspace{-0.5em}
  \label{Fig:AblationRD}
\end{figure}

{
\begin{table}[t]
  \centering
  \caption{BD-Rate (\%) and BD-PSNR (dB) gains of the proposed modules.} 
  \fontsize{8}{10}\selectfont
  \setlength\tabcolsep{7pt}
  \vspace{-0.5em}
\begin{tabular}{lrrrrrrrr}
\toprule
    & \multicolumn{4}{c}{Compared with Baseline} & \multicolumn{4}{c}{Compared with G-PCC octree} \\
Methods & \multicolumn{2}{c}{BD-Rate} & \multicolumn{2}{c}{BD-PSNR} & \multicolumn{2}{c}{BD-Rate} & \multicolumn{2}{c}{BD-PSNR} \\
    & D1  & D2  & D1  & D2  & D1  & D2  & D1  & D2 \\
\midrule
Baseline & 0.00 & 0.00 & 0.00 & 0.00 & -10.04 & -10.08 & 1.13 & 1.14 \\
 + GRED & -3.78 & -3.78 & 0.40 & 0.40 & -13.44 & -13.47 & 1.49 & 1.50 \\
 + GRED + XFP & -13.22 & -13.21 & 1.46 & 1.46 & -21.93 & -21.95 & 2.53 & 2.54 \\
\bottomrule
\end{tabular}%
  \label{Tab:Ablation}%
  \vspace{-0.5em}
\end{table}%
}

\subsection{Ablation Studies}
To evaluate the individual contributions of the proposed components, we conduct ablation studies on the GRED module and the XFP module. Each component is systematically removed to assess its impact on overall RD performance. 


\textbf{Ablation of XFP}. We ablate the XFP module by removing cross-scale feature propagation. As shown by the ``Baseline + GRED'' curves in Fig.~\ref{Fig:AblationRD}, this leads to a clear RD drop at high quantization precisions, but has negligible impact at low precisions, consistent with our theoretical analysis. Quantitative results in Table~\ref{Tab:Ablation} show that this removal results in a degradation of approximately 1.06 dB, indicating that the integration of cross-scale information through XFP is critical for improving RD performance. 

\textbf{Ablation of GRED}. To evaluate the effectiveness of the GRED module, we removed GRED on top of the XFP ablation. In this setting, the dense features extracted from the shallow level are no longer utilized for predicting the occupancy of deeper levels. 
As a result, the model is directly exposed to the HRCS problem under high-resolution encoding. 
The corresponding RD performance is shown as the ``Baseline'' curves in Fig.~\ref{Fig:AblationRD}. 
The results reveal that, as the resolution increases, the RD performance of the model without GRED declines significantly compared to the variant equipped with GRED.  
Quantitative results in Table~\ref{Tab:Ablation} show that removing the GRED module leads to a further performance drop of approximately 0.40 dB. This performance gap highlights the positive impact of the GRED module in mitigating the effects of HRCS and enhancing performance.

\section{Conclusion}
This paper proposes a cross-platform, real-time neural compression framework for LiDAR point clouds, jointly addressing deterministic coding requirements and the High-Resolution Contextual Sparsity challenge. 
We introduce an octree-based design that integrates the Geometry Re-Densification module and the Cross-Scale Feature Propagation module, enabling effective intra-scale and cross-scale context modeling under extreme sparsity. In addition, an integer-only inference pipeline is developed to ensure bit-exact coding across heterogeneous platforms. Extensive experiments demonstrate that the proposed framework achieves competitive rate-distortion performance while maintaining real-time encoding and decoding speed. 



\section*{Acknowledgements}
This work was supported in part by the National Natural Science Foundation of China and the Major Key Project of PCL under Grant No. PCL2024A04.

%
%
\bibliographystyle{splncs04}
\bibliography{reference}
\end{document}


\title{Supplementary Material: Towards Practical Lossless Neural Compression for LiDAR Point Clouds} 


\author{First Author\inst{1}\orcidlink{0000-1111-2222-3333} \and
Second Author\inst{2,3}\orcidlink{1111-2222-3333-4444} \and
Third Author\inst{3}\orcidlink{2222-3333-4444-5555}}

\authorrunning{F.~Author et al.}

\institute{Princeton University, Princeton NJ 08544, USA \and
Springer Heidelberg, Tiergartenstr.~17, 69121 Heidelberg, Germany
\email{lncs@springer.com}\\
\url{http://www.springer.com/gp/computer-science/lncs} \and
ABC Institute, Rupert-Karls-University Heidelberg, Heidelberg, Germany\\
\email{\{abc,lncs\}@uni-heidelberg.de}}

\maketitle

\section{Differences with Existing Methods}
\label{Sec:DifferenceswithExistingMethods}
This work explicitly identifies and addresses the High-Resolution Contextual Sparsity (HRCS) problem in point cloud compression. 
While it is difficult to determine whether existing methods implicitly mitigate HRCS in a generalizable way, we examine several representative approaches to clarify this issue.

\textbf{Transformer-based Methods (\eg, EHEM, OctAttention).} These methods represent octree nodes as explicit feature vectors and use transformer architectures to capture long-range dependencies between nodes, thereby enlarging the receptive field. However, such methods essentially process octree nodes in a 1D sequence space rather than in the native 3D space. 
While their attention mechanism can implicitly model 3D geometry, it does not explicitly preserve 3D structural information and instead relies heavily on learned embeddings. As a result, these methods typically require large attention windows (e.g., 8192 nodes in Light EHEM) to achieve competitive performance. 
According to our analysis, this leads to about 10× the FLOPs of our approach. Thus, while transformer-based methods might sidestep the HRCS problem by modeling 1D node sequences via long-range attention, this comes at the cost of significantly increased computational complexity and latency. 

\textbf{Sparse Convolution-based Methods (\eg, Unicorn, SparsePCGC).} These methods exploit voxel-level neighborhoods via sparse convolutions in a coarse-to-fine reconstruction pipeline. By design, sparse convolutions skip computation on empty voxels to improve efficiency. However, this sparsity impedes information propagation across voxels when the point cloud is highly sparse at finer scales.
For instance, if an occupied voxel is surrounded by 26 empty neighbors ($3\times3\times3-1$), no amount of stacking sparse convolution layers with a $3\times3\times3$ kernel can retrieve geometric context for this voxel, making it isolated. Although SparsePCGC alleviates this problem to some extent by increasing the depth of the convolution blocks, 
it suffers from high coding latency. 

\begin{figure}[t]
  \centering
  \includegraphics[scale=0.7,trim=0 0 0 0,clip]{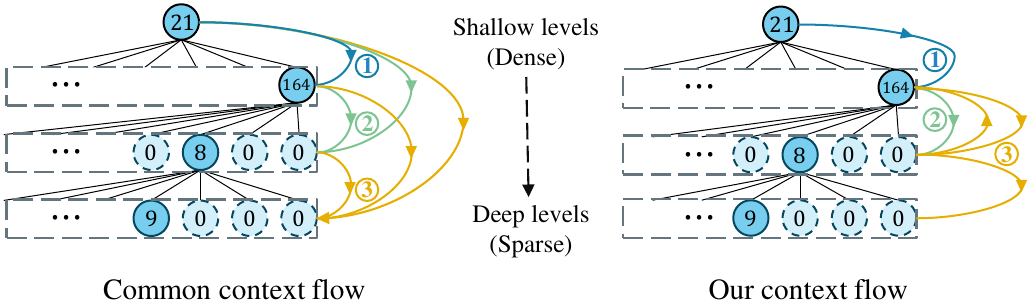}
  \caption{Comparison of context flows in existing octree-based methods (left) and our approach (right). Lines with different colors and numbers indicate different feature extraction and occupancy prediction steps. In common designs, context flow is typically unidirectional, and encoding/decoding at each octree level depends on geometry from multiple preceding levels. By contrast, our approach adopts a bidirectional context flow at deeper octree levels, while at shallow levels, each octree level relies only on the feature and geometry of the immediately preceding level. }
  \label{Fig:ContextFlow}
\end{figure}

In summary, while existing methods may touch on related ideas, none have explicitly recognized HRCS as a distinct challenge or introduced a targeted practical solution for it. 
For clarity, we provide a visualization comparing the context modeling processes of existing methods with our approach in Fig.~\ref{Fig:ContextFlow}, highlighting the key differences in design.

\section{Detailed Model Structure}
\begin{figure*}[t]
  \centering
  \includegraphics[width=\linewidth]{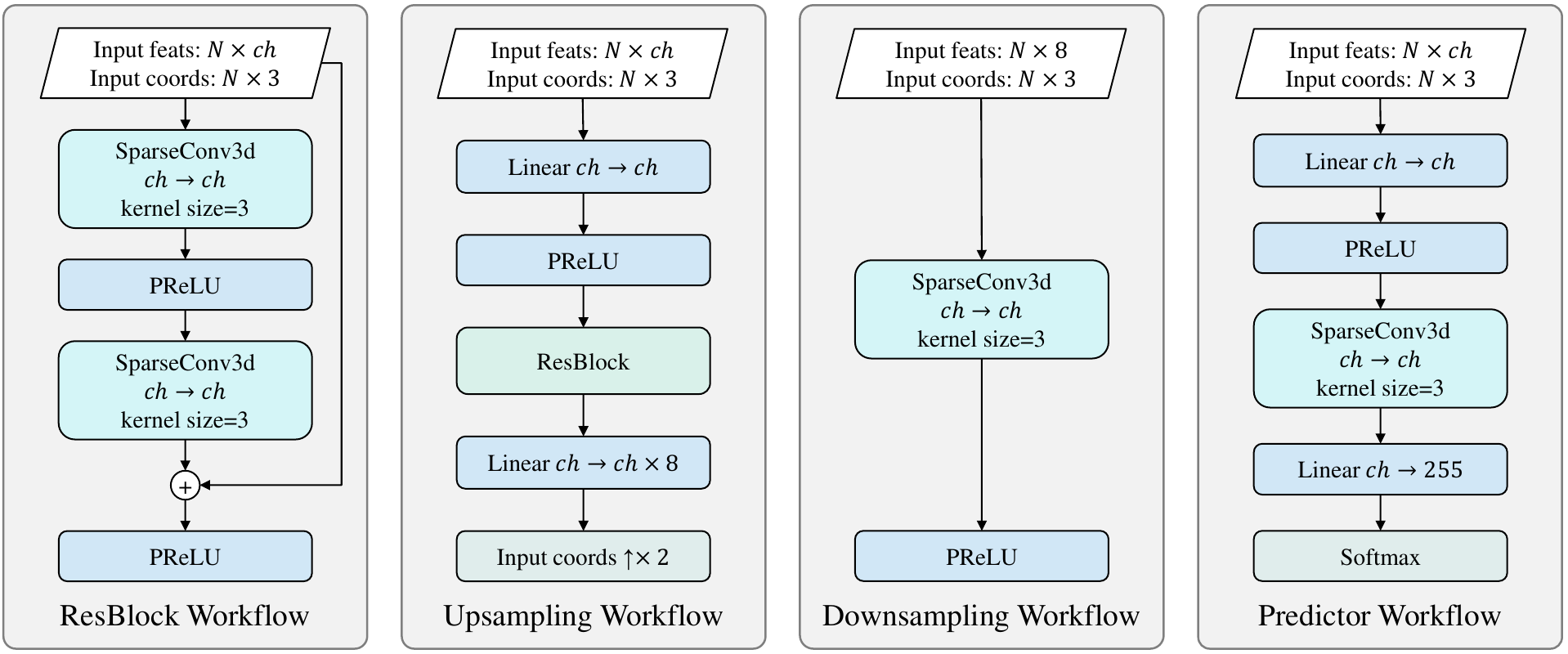}
  \caption{Illustration of the detailed workflows of the $\mathrm{Upsampling}(\cdot)$, $\mathrm{ResBlock}(\cdot)$, $\mathrm{Downsampling}(\cdot)$, and $\mathrm{Predictor}(\cdot)$ modules within the proposed network architecture.}
  \label{Fig:AppendixStructure}
\end{figure*}

To clarify the structure and workflow of the key modules in our network, we present a detailed breakdown of the $\mathrm{Upsampling}(\cdot)$, $\mathrm{ResBlock}(\cdot)$, $\mathrm{Downsampling}(\cdot)$, and $\mathrm{Predictor}(\cdot)$ components. 
The detailed workflow is illustrated in Fig.~\ref{Fig:AppendixStructure}.

In the current implementation, our primary goal is to validate the core ideas of geometry re-densification and cross-scale feature propagation. To ensure an isolated evaluation of the proposed modules, we intentionally omit the cross-scale parameter-sharing mechanisms employed in previous works. 
Consequently, our model is less parameter-efficient compared to prior methods, as shown in Table~\ref{Tab:AppendixParam}, 
since the number of parameters grows linearly with the number of explicitly modeled octree levels (\ie, $L-11$ levels in this paper).
For the geometry remaining at the maximum downsampling level, we directly encode the coordinates based on their symbol frequencies.
In future work, we plan to improve parameter efficiency by introducing level-aware neural blocks that support parameter sharing across octree levels.

\section{Implementation Details}
\label{Sec:ImplementationDetails}
This section outlines the details of our implementation, including quantization strategies for the KITTI dataset, octree-based operations, training configurations, and evaluation metrics.

\begin{table}[tb]
  \centering
  \setlength\tabcolsep{3pt}
  \caption{Comparison of the number of model parameters.}
  \footnotesize
    \begin{tabular}{llllll}
    \toprule
    Methods & PCGCv2 & OctAttn & EHEM & Ours & RENO \\
    \midrule
    \#Parameters & 0.77M & 6.99M & 13.01M & 102.51M & 0.28M \\
    \bottomrule
    \end{tabular}%
  \label{Tab:AppendixParam}%
\end{table}%

\subsection{Quantization of KITTI Dataset}
The KITTI point clouds are not officially quantized, and two different quantization approaches have therefore been adopted in prior works:
\begin{enumerate}
    \item The first approach normalizes the point clouds within a bounding box of size $400 \times 400 \times 400$ centered at the origin $(0, 0, 0)$, scales the coordinates by $2^{16}$, and then applies quantization.
    \item The second approach, adopted by RENO, scales the original floating-point coordinates by 10000, followed by quantization using an additional scale factor $\mathrm{posQ} \in \{8, 16, 32, 64, 128, 256, 512\}$ to generate point clouds of different precision levels.
\end{enumerate}

These two approaches yield point clouds of different fidelity, leading to different PSNR values even under lossless compression settings, where the only distortion arises from the quantization process. Consequently, 
the RD points of RENO do not align with those of other methods, despite all of them being lossless codecs.
While it is technically feasible to unify the quantization strategy, we follow RENO’s official setting to report its results, as it better reflects the RD performance under its original setting.

For clarity, we refer to the highest-precision results from both quantization approaches as \emph{16-bit quantization} throughout this paper. In our experiments, the octree level is varied from 11 to 16, enabling control over the RD trade-off by adjusting the spatial resolution.

\subsection{Implementation of Octree Operations}

Two components are essential for implementing octree-based operations with sparse convolution: coordinate upsampling/downsampling and occupancy code generation.

\textbf{Coordinate Sampling}.
To efficiently generate node coordinates for all octree levels, we first sort the input coordinates in Morton order, which exploits the hierarchical spatial locality of the octree.
Starting from the input coordinates, we repeatedly divide them by 2, apply floor rounding, and remove consecutive duplicates. This yields the coordinates of nodes at progressively shallower octree levels. 
For coordinate upsampling, we reconstruct child node coordinates by adding a pre-defined offset matrix (leveraging matrix broadcasting) to the parent coordinates. We then apply a masking operation to discard coordinates corresponding to unoccupied nodes.
Importantly, this process preserves the original Morton order of the coordinate matrix, ensuring consistent alignment between the encoder and decoder without the need for explicit reordering.

\textbf{Occupancy Code Generation}.
To generate the 0–255 occupancy codes, we apply a fixed-weight sparse convolution with kernel size 2 and stride 2, using an all-one input feature tensor.
Each 8-neighbor group (in $2 \times 2 \times 2$) is encoded as an 8-bit occupancy code.
For the reverse process, the occupancy code can be efficiently decoded into binary masks using bitwise operations and matrix broadcasting.

\subsection{Training}
\textbf{Loss Function}. To train the proposed model, we adopt the standard cross-entropy loss, which is widely used in octree-based PCC. Specifically, the model outputs a probability distribution $\mathbf{Q} \in \mathbb{R}^{N \times 255}$, 
where each row corresponds to the predicted occupancy distribution of one of the $N$ nodes over the 255 possible occupancy codes. Let the ground truth occupancy codes be represented by $\mathbf{X} \in \{1, \dots, 255\}^N$. The loss function is defined as:
\begin{equation}
    \mathcal{L}_{CE} = -\frac{1}{N} \sum_{i=1}^{N} \log \mathbf{Q}_{i, \mathbf{X}_i},
\end{equation}
where $\mathbf{Q}_{i, \mathbf{X}_i}$ denotes the predicted probability for the ground truth occupancy code $\mathbf{X}_i$ at the $i$-th node. This loss encourages the model to assign higher probabilities to the correct occupancy codes.

\textbf{Other Settings}. We adopt the AdamW optimizer 
with a weight decay of 0.0001 and a learning rate of 0.0001. Gradient clipping is applied with a maximum norm threshold of 1.0 to stabilize training.
The model is trained for 60 epochs with a batch size of 8. 
All experiments were conducted on a computer equipped with an AMD EPYC 7R32 CPU and $2\times$ 4090 GPUs. Training takes approximately 4 days.

\section{More Quantitative Analysis}
\label{Sec:MoreQuantitativeAnalysis}
In this section, we provide further quantitative analysis to complement the main results.

\subsection{Ablation on the Choice of $t$}
\begin{figure}[tb]
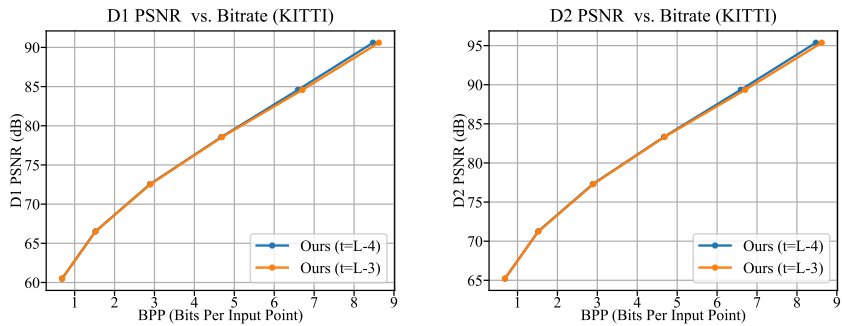

  \centering
  \includegraphics[width=0.43\linewidth,trim=5 6 5 5,clip]{data/KITTI Ablation t/sample-wise D1 PSNR (dB)/D1 PSNR (dB) KITTI.pdf}
  \hfil
  \includegraphics[width=0.43\linewidth,trim=5 6 5 5,clip]{data/KITTI Ablation t/sample-wise D2 PSNR (dB)/D2 PSNR (dB) KITTI.pdf}
  \caption{Rate-distortion performance comparison for different choices of $t$ on KITTI.}
  \label{Fig:AppendixAblationtRD}
\end{figure}

As demonstrated in the Method section, we apply the re-densification module only at levels $l > t$, making $t$ a key hyperparameter. A smaller $t$ applies re-densification to more levels, enabling more efficient context modeling but at the cost of higher re-densification overhead. Moreover, applying re-densification to shallower levels with dense geometry is often redundant. Therefore, the marginal benefit of reducing $t$ quickly diminishes. 
In our experiments, unless otherwise stated, we set $t = L - 4$ by default, where $L$ is the deepest octree level. Here, we compare the performance of two settings: $t = L - 4$ and $t = L - 3$. The resulting RD performance is shown in Fig.~\ref{Fig:AppendixAblationtRD} and Table~\ref{Tab:AppendixAblationtRD}.
We observe slight improvements at high precision when reducing $t$ from $L - 3$ to $L - 4$. However, the average gain across all bitrates is relatively minor. Overall, the setting $t = L - 4$ achieves a 0.76\% bitrate reduction on the KITTI dataset.

\begin{table}[tb]
  \centering
  \setlength\tabcolsep{10pt}
  \caption{BD gains of the $t=L-4$ setting over the $t=L-3$ setting.}
  \footnotesize
    \begin{tabular}{lrrrr}
    \toprule
         & \multicolumn{2}{c}{BD-Rate (\%)} & \multicolumn{2}{c}{BD-PSNR (dB)} \\
        Dataset & \multicolumn{1}{c}{D1} & \multicolumn{1}{c}{D2} & \multicolumn{1}{c}{D1} & \multicolumn{1}{c}{D2} \\
    \midrule
    KITTI & -0.760 & -0.759 & 0.093 & 0.093 \\
    \bottomrule
    \end{tabular}%
  \label{Tab:AppendixAblationtRD}%
\end{table}%

{
\begin{table}[htb]
  \centering
  \caption{Detailed complexity metrics of our model on the KITTI dataset.}
  \fontsize{8}{10}\selectfont
  \setlength\tabcolsep{7pt}
    \begin{tabular}{cccrr}
    \toprule
Prec. (bits) & Enc Mem (GB) & Dec Mem (GB) & Enc GFLOPs & Dec GFLOPs \\
\midrule
16 & 2.1 & 2.0 & 752.9 & 752.9 \\
15 & 1.6 & 1.5 & 455.7 & 455.7 \\
14 & 1.2 & 1.1 & 238.2 & 238.2 \\
13 & 0.9 & 0.8 & 108.9 & 108.9 \\
12 & 0.7 & 0.6 & 44.7 & 44.7 \\
11 & 0.6 & 0.6 & 16.9 & 16.9 \\
    \bottomrule
    \end{tabular}%
  \label{Tab:AppendixMoreComplexityResultsOurs}%
\end{table}%
}

{
\begin{table*}[htb]
  \centering
  \caption{Detailed complexity metrics of Light EHEM on the KITTI dataset.}
  \fontsize{8}{13}\selectfont
  \setlength\tabcolsep{11pt}
    \begin{tabular}{cccrr}
    \toprule
        & Enc/Dec & GFLOPs & \multicolumn{1}{c}{Number of} & \multicolumn{1}{c}{Enc/Dec}    \\
    Prec. (bits) & Mem (GB) & per Window & \multicolumn{1}{c}{Octree Nodes} & \multicolumn{1}{c}{GFLOPs} \\
    \midrule
    16  & 2.6 & 102.9 & 421911.3 & 5299.6 \\
    15  & 2.6 & 102.9 & 302393.4 & 3798.4 \\
    14  & 2.6 & 102.9 & 191616.0 & 2406.9 \\
    13  & 2.6 & 102.9 & 105002.2 & 1318.9 \\
    12  & 2.6 & 102.9 & 49486.4 & 621.6  \\
    11  & 2.6 & 102.9 & 20591.8 & 258.7 \\
    \bottomrule
    \end{tabular}%
  \label{Tab:AppendixMoreComplexityResultsLightEHEM}%
\end{table*}%
}

\subsection{FLOPs}
\label{Sec:MoreComputationalEfficiency}
We conducted additional experiments to measure the computational complexity of encoding and decoding, and the results are summarized in Table~\ref{Tab:AppendixMoreComplexityResultsOurs}. Note that the FLOPs of sparse convolution depend on the sparsity of the input point clouds, which varies across samples. 
The reported FLOPs of our model are averaged over the KITTI test set.
For comparison, we provide the official metrics of Light EHEM in Table~\ref{Tab:AppendixMoreComplexityResultsLightEHEM}. Note that the FLOPs reported by Light EHEM are measured per window (with 8192 octree nodes). To ensure a fair comparison, 
we converted this into an expected average per sample using the mean octree node count of the quantized KITTI point clouds. 
These results show that our approach is more efficient than the transformer-based Light EHEM in terms of memory usage, FLOPs, and execution speed.

\subsection{Sequence-wise Performance}
\begin{figure*}
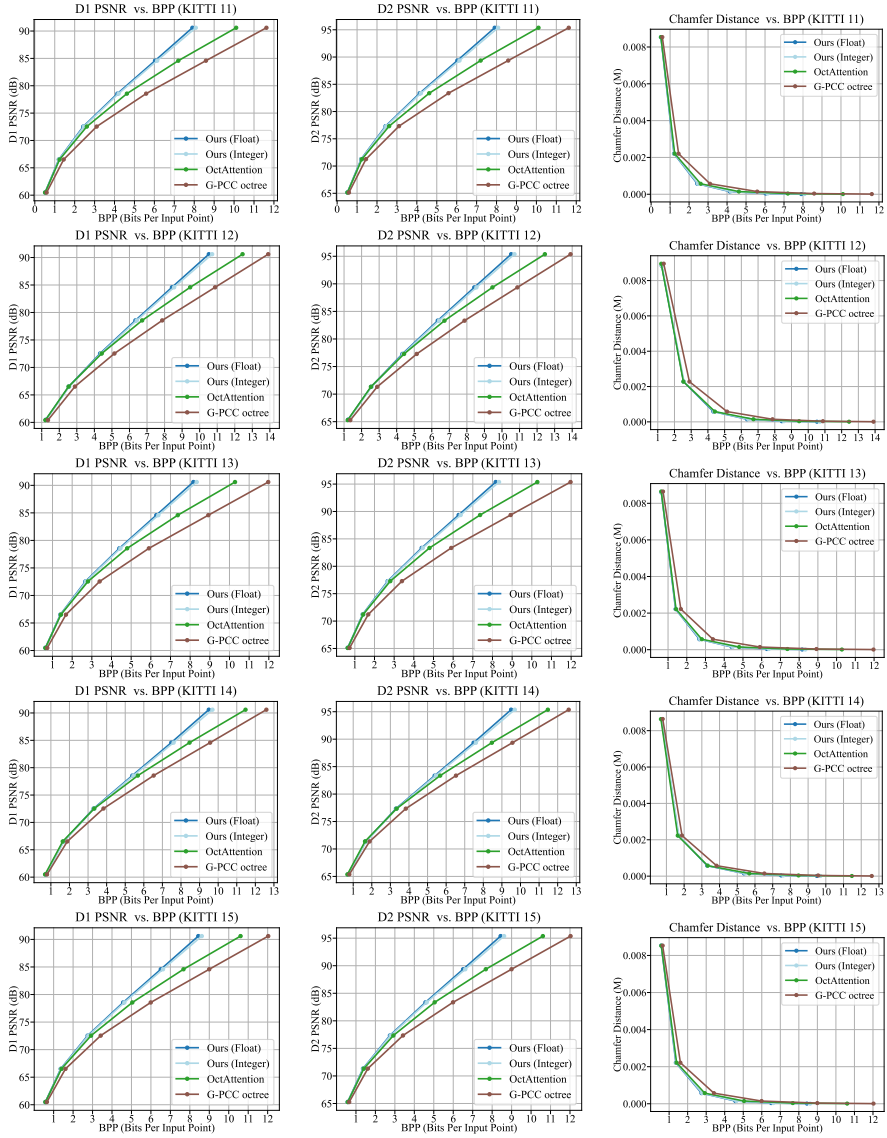

  \centering
  \includegraphics[width=0.3\linewidth,trim=5 6 5 5,clip]{data/KITTI sequence-wise/sample-wise D1 PSNR (dB)/D1 PSNR (dB) KITTI 11.pdf}
  \hfil
  \includegraphics[width=0.3\linewidth,trim=5 6 5 5,clip]{data/KITTI sequence-wise/sample-wise D2 PSNR (dB)/D2 PSNR (dB) KITTI 11.pdf}
  \hfil
  \includegraphics[width=0.3\linewidth,trim=5 6 5 5,clip]{data/KITTI sequence-wise/sample-wise Chamfer Distance (M)/Chamfer Distance (M) KITTI 11.pdf}\\
  \includegraphics[width=0.3\linewidth,trim=5 6 5 5,clip]{data/KITTI sequence-wise/sample-wise D1 PSNR (dB)/D1 PSNR (dB) KITTI 12.pdf}
  \hfil
  \includegraphics[width=0.3\linewidth,trim=5 6 5 5,clip]{data/KITTI sequence-wise/sample-wise D2 PSNR (dB)/D2 PSNR (dB) KITTI 12.pdf}
  \hfil
  \includegraphics[width=0.3\linewidth,trim=5 6 5 5,clip]{data/KITTI sequence-wise/sample-wise Chamfer Distance (M)/Chamfer Distance (M) KITTI 12.pdf}\\
  \includegraphics[width=0.3\linewidth,trim=5 6 5 5,clip]{data/KITTI sequence-wise/sample-wise D1 PSNR (dB)/D1 PSNR (dB) KITTI 13.pdf}
  \hfil
  \includegraphics[width=0.3\linewidth,trim=5 6 5 5,clip]{data/KITTI sequence-wise/sample-wise D2 PSNR (dB)/D2 PSNR (dB) KITTI 13.pdf}
  \hfil
  \includegraphics[width=0.3\linewidth,trim=5 6 5 5,clip]{data/KITTI sequence-wise/sample-wise Chamfer Distance (M)/Chamfer Distance (M) KITTI 13.pdf}\\
  \includegraphics[width=0.3\linewidth,trim=5 6 5 5,clip]{data/KITTI sequence-wise/sample-wise D1 PSNR (dB)/D1 PSNR (dB) KITTI 14.pdf}
  \hfil
  \includegraphics[width=0.3\linewidth,trim=5 6 5 5,clip]{data/KITTI sequence-wise/sample-wise D2 PSNR (dB)/D2 PSNR (dB) KITTI 14.pdf}
  \hfil
  \includegraphics[width=0.3\linewidth,trim=5 6 5 5,clip]{data/KITTI sequence-wise/sample-wise Chamfer Distance (M)/Chamfer Distance (M) KITTI 14.pdf}\\
  \includegraphics[width=0.3\linewidth,trim=5 6 5 5,clip]{data/KITTI sequence-wise/sample-wise D1 PSNR (dB)/D1 PSNR (dB) KITTI 15.pdf}
  \hfil
  \includegraphics[width=0.3\linewidth,trim=5 6 5 5,clip]{data/KITTI sequence-wise/sample-wise D2 PSNR (dB)/D2 PSNR (dB) KITTI 15.pdf}
  \hfil
  \includegraphics[width=0.3\linewidth,trim=5 6 5 5,clip]{data/KITTI sequence-wise/sample-wise Chamfer Distance (M)/Chamfer Distance (M) KITTI 15.pdf}
  \caption{Sequence-wise rate-distortion performance comparison on the KITTI test sequences $\#11$-$\#15$.}
  \label{Fig:AppendixRD1}
\end{figure*}

\begin{figure*}
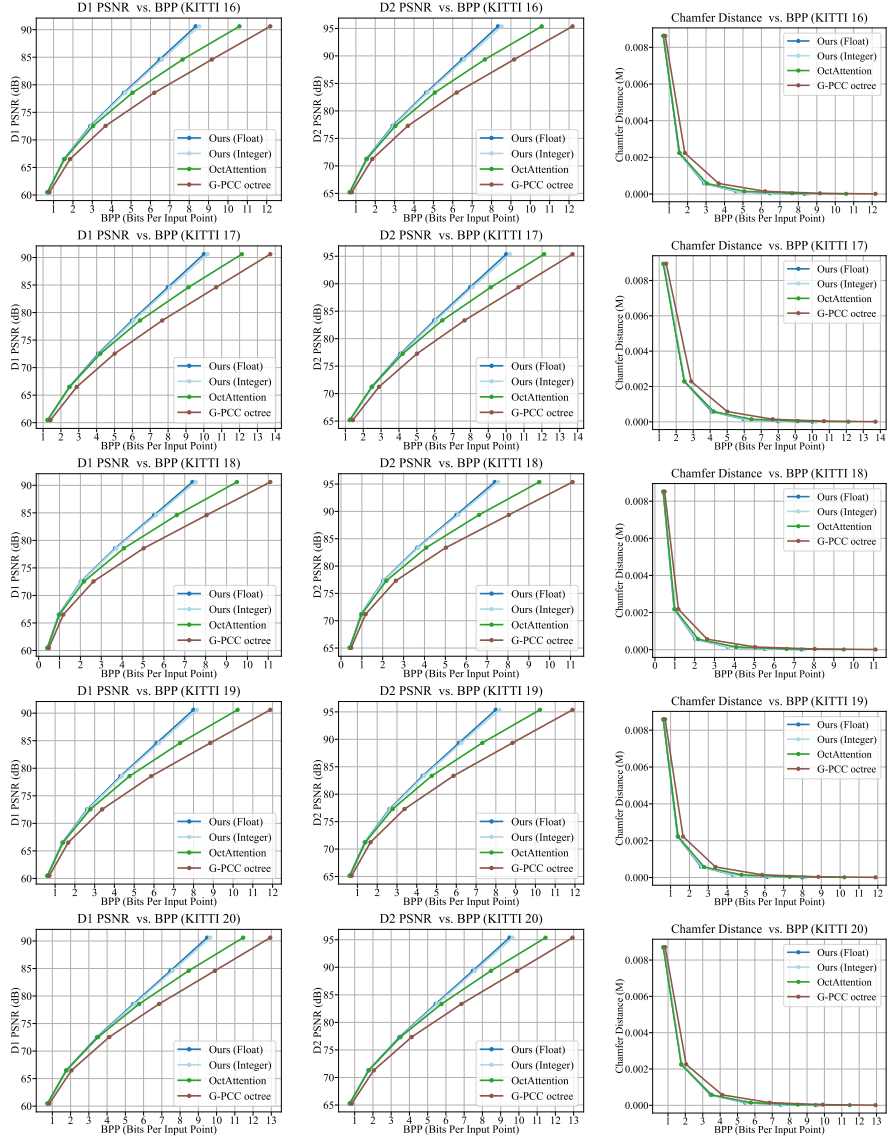

  \centering
  \includegraphics[width=0.3\linewidth,trim=5 6 5 5,clip]{data/KITTI sequence-wise/sample-wise D1 PSNR (dB)/D1 PSNR (dB) KITTI 16.pdf}
  \hfil
  \includegraphics[width=0.3\linewidth,trim=5 6 5 5,clip]{data/KITTI sequence-wise/sample-wise D2 PSNR (dB)/D2 PSNR (dB) KITTI 16.pdf}
  \hfil
  \includegraphics[width=0.3\linewidth,trim=5 6 5 5,clip]{data/KITTI sequence-wise/sample-wise Chamfer Distance (M)/Chamfer Distance (M) KITTI 16.pdf}\\
  \includegraphics[width=0.3\linewidth,trim=5 6 5 5,clip]{data/KITTI sequence-wise/sample-wise D1 PSNR (dB)/D1 PSNR (dB) KITTI 17.pdf}
  \hfil
  \includegraphics[width=0.3\linewidth,trim=5 6 5 5,clip]{data/KITTI sequence-wise/sample-wise D2 PSNR (dB)/D2 PSNR (dB) KITTI 17.pdf}
  \hfil
  \includegraphics[width=0.3\linewidth,trim=5 6 5 5,clip]{data/KITTI sequence-wise/sample-wise Chamfer Distance (M)/Chamfer Distance (M) KITTI 17.pdf}\\
  \includegraphics[width=0.3\linewidth,trim=5 6 5 5,clip]{data/KITTI sequence-wise/sample-wise D1 PSNR (dB)/D1 PSNR (dB) KITTI 18.pdf}
  \hfil
  \includegraphics[width=0.3\linewidth,trim=5 6 5 5,clip]{data/KITTI sequence-wise/sample-wise D2 PSNR (dB)/D2 PSNR (dB) KITTI 18.pdf}
  \hfil
  \includegraphics[width=0.3\linewidth,trim=5 6 5 5,clip]{data/KITTI sequence-wise/sample-wise Chamfer Distance (M)/Chamfer Distance (M) KITTI 18.pdf}\\
  \includegraphics[width=0.3\linewidth,trim=5 6 5 5,clip]{data/KITTI sequence-wise/sample-wise D1 PSNR (dB)/D1 PSNR (dB) KITTI 19.pdf}
  \hfil
  \includegraphics[width=0.3\linewidth,trim=5 6 5 5,clip]{data/KITTI sequence-wise/sample-wise D2 PSNR (dB)/D2 PSNR (dB) KITTI 19.pdf}
  \hfil
  \includegraphics[width=0.3\linewidth,trim=5 6 5 5,clip]{data/KITTI sequence-wise/sample-wise Chamfer Distance (M)/Chamfer Distance (M) KITTI 19.pdf}\\
  \includegraphics[width=0.3\linewidth,trim=5 6 5 5,clip]{data/KITTI sequence-wise/sample-wise D1 PSNR (dB)/D1 PSNR (dB) KITTI 20.pdf}
  \hfil
  \includegraphics[width=0.3\linewidth,trim=5 6 5 5,clip]{data/KITTI sequence-wise/sample-wise D2 PSNR (dB)/D2 PSNR (dB) KITTI 20.pdf}
  \hfil
  \includegraphics[width=0.3\linewidth,trim=5 6 5 5,clip]{data/KITTI sequence-wise/sample-wise Chamfer Distance (M)/Chamfer Distance (M) KITTI 20.pdf}\\
  \caption{Sequence-wise rate-distortion performance comparison on the KITTI test sequences $\#16$-$\#20$.}
  \label{Fig:AppendixRD2}
\end{figure*}

To account for the variation in sequence characteristics within the KITTI dataset, we provide sequence-wise RD performance for further analysis and comparison among G-PCC, OctAttention, and our method.
The evaluation metrics include D1 PSNR, D2 PSNR, and Chamfer Distance (CD), a widely adopted geometric distortion metric. The results are illustrated in Fig.~\ref{Fig:AppendixRD1} (sequences $\#11$-$\#15$) and Fig.~\ref{Fig:AppendixRD2} (sequences $\#16$-$\#20$). 
These plots highlight the consistency and robustness of our method across different scenes.

\section{Qualitative Analysis}
\begin{figure*}
  \centering  
  \subfloat[Ours: 72.52dB (D1), 2.46BPP]{\includegraphics[width=0.5\linewidth,trim=0 0 0 0,clip]{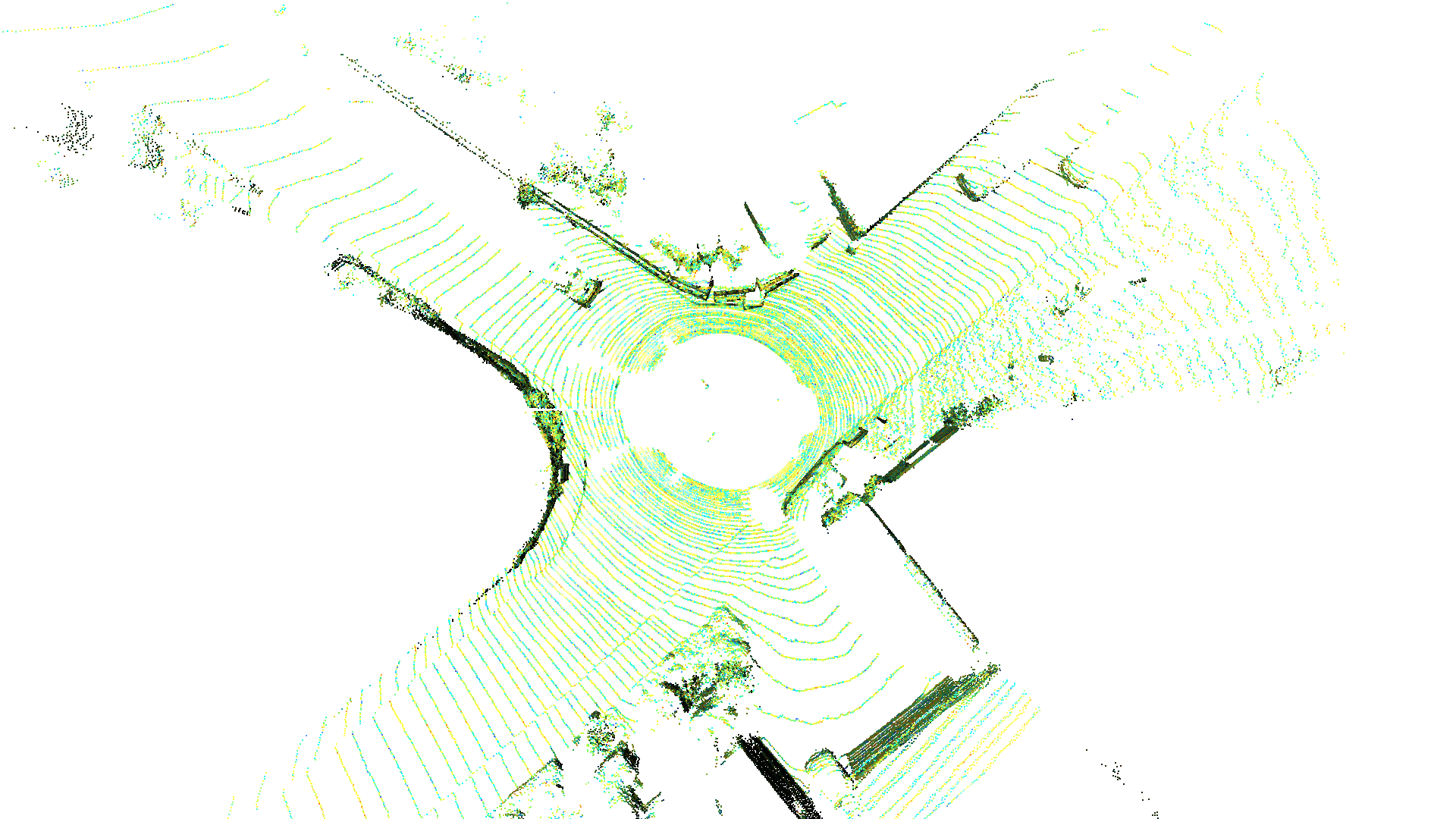}}
  \hfil
  \subfloat[RENO: 70.17dB (D1), 2.26BPP]{\includegraphics[width=0.5\linewidth,trim=0 0 0 0,clip]{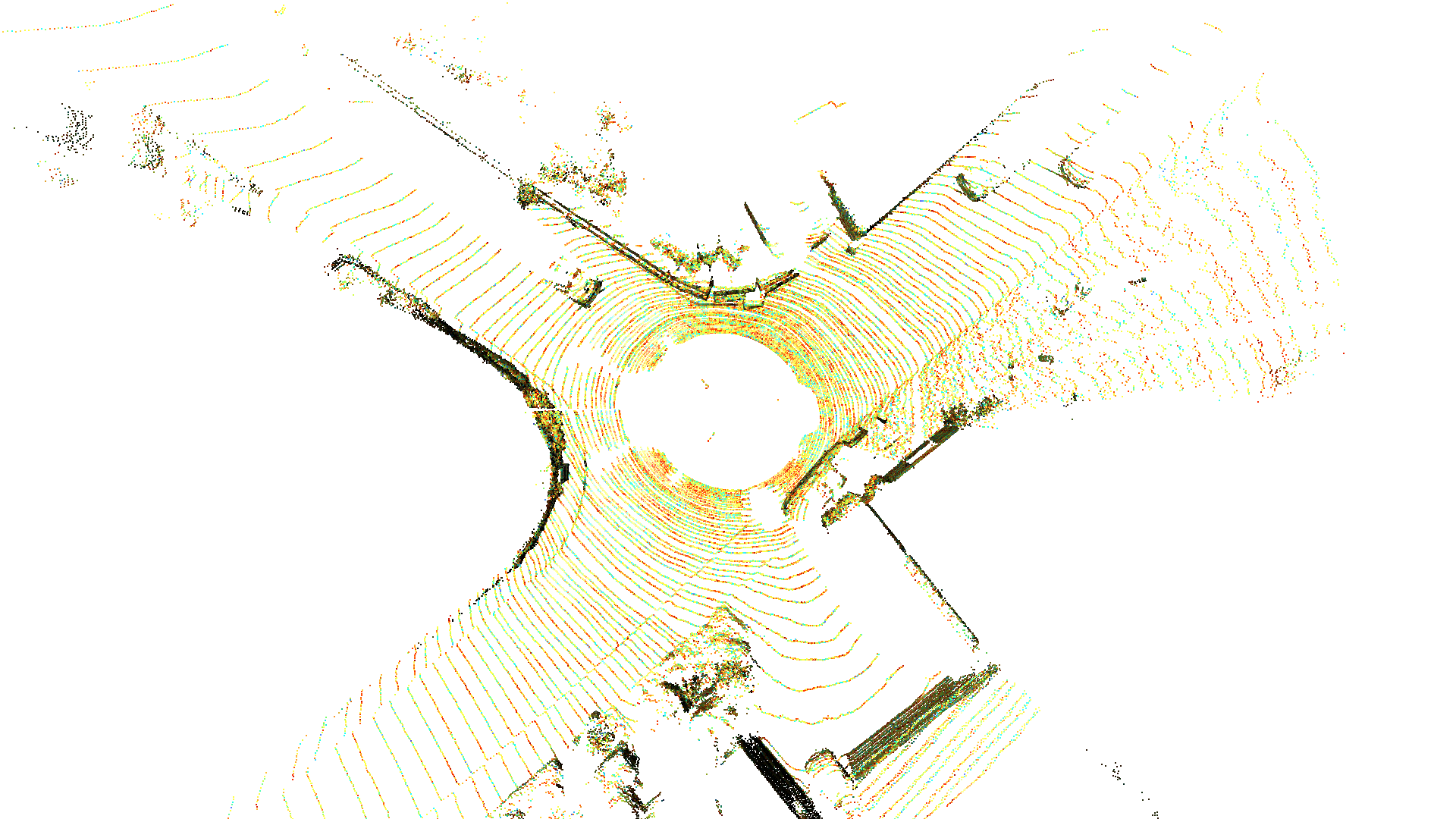}} \\
  \vspace{1em}
  \subfloat[Ours: 78.55dB (D1), 4.11BPP]{\includegraphics[width=0.5\linewidth,trim=0 0 0 0,clip]{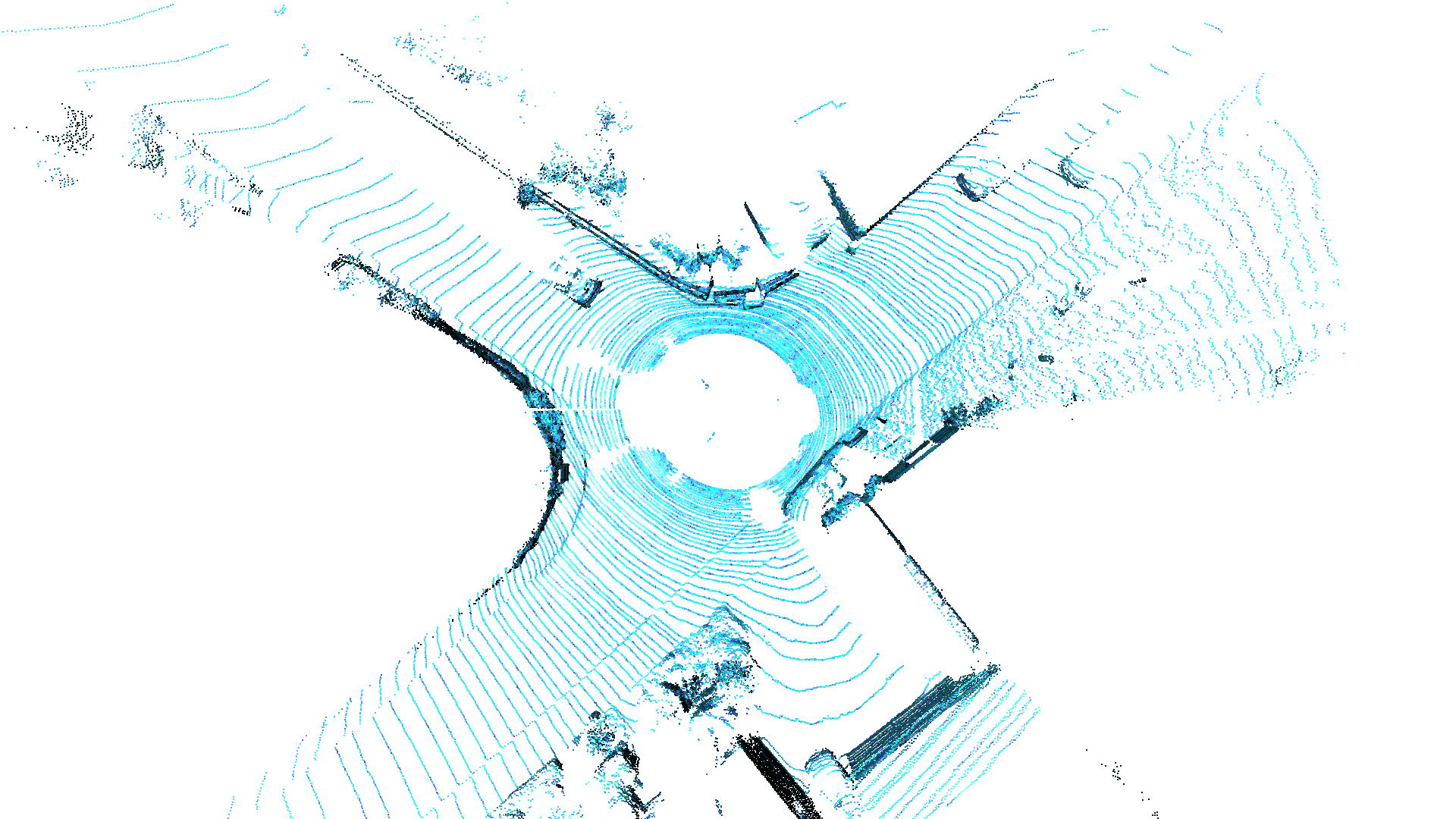}}
  \hfil
  \subfloat[RENO: 76.21dB (D1), 4.17BPP]{\includegraphics[width=0.5\linewidth,trim=0 0 0 0,clip]{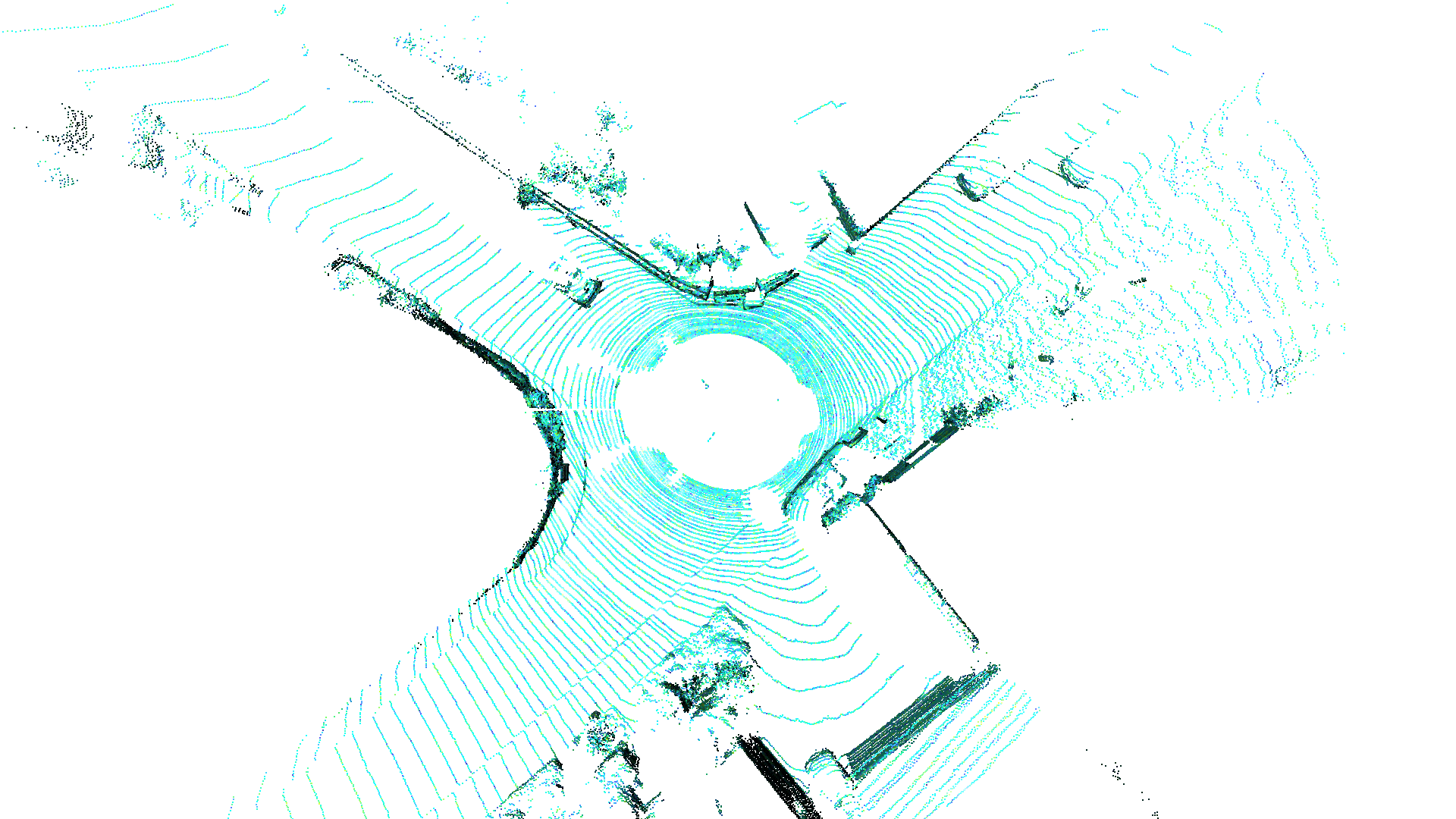}} \\
  \vspace{1em}
  \subfloat[Ours: 84.59dB (D1), 5.90BPP]{\includegraphics[width=0.5\linewidth,trim=0 0 0 0,clip]{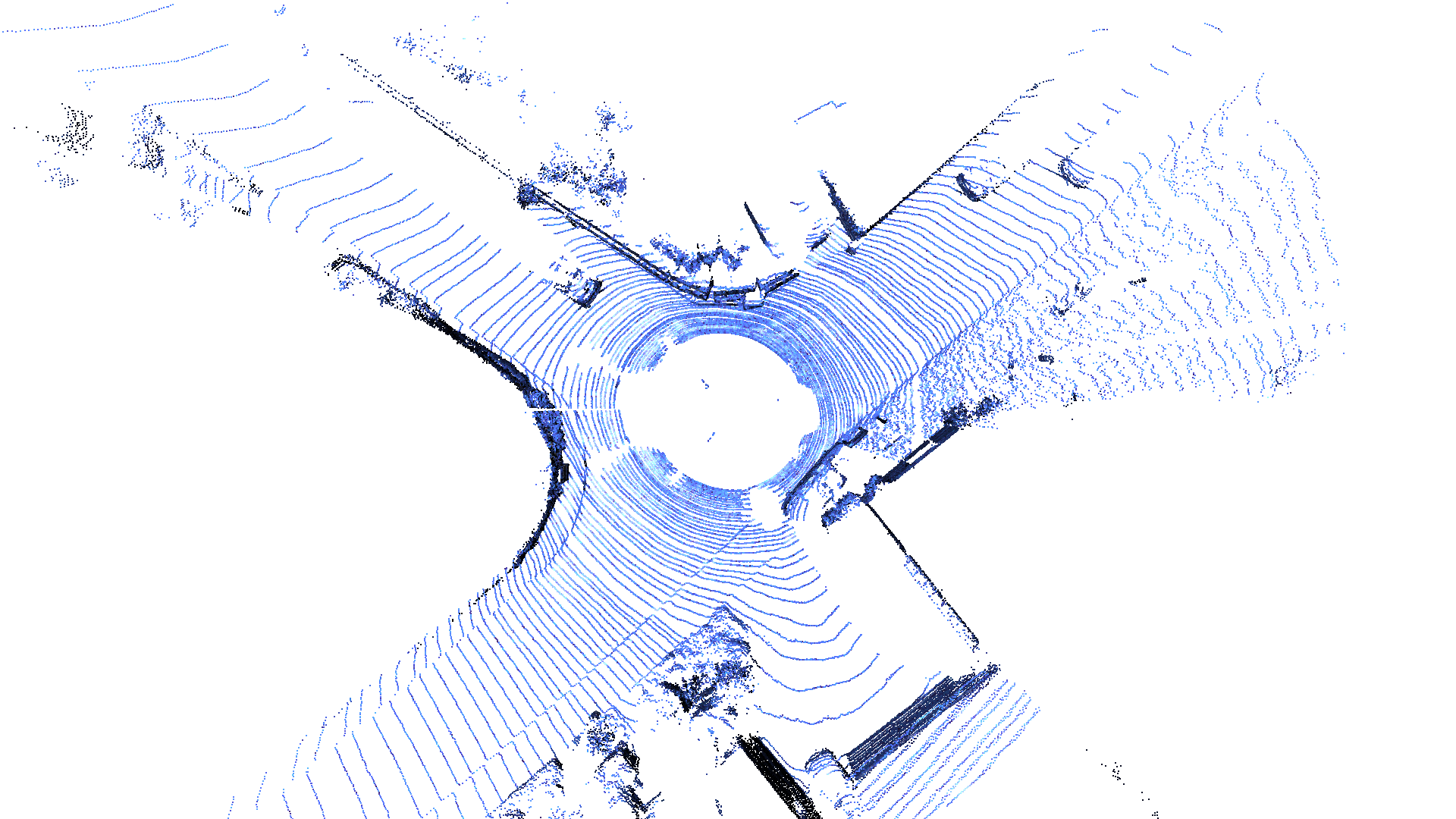}}
  \hfil
  \subfloat[RENO: 82.25dB (D1), 6.53BPP]{\includegraphics[width=0.5\linewidth,trim=0 0 0 0,clip]{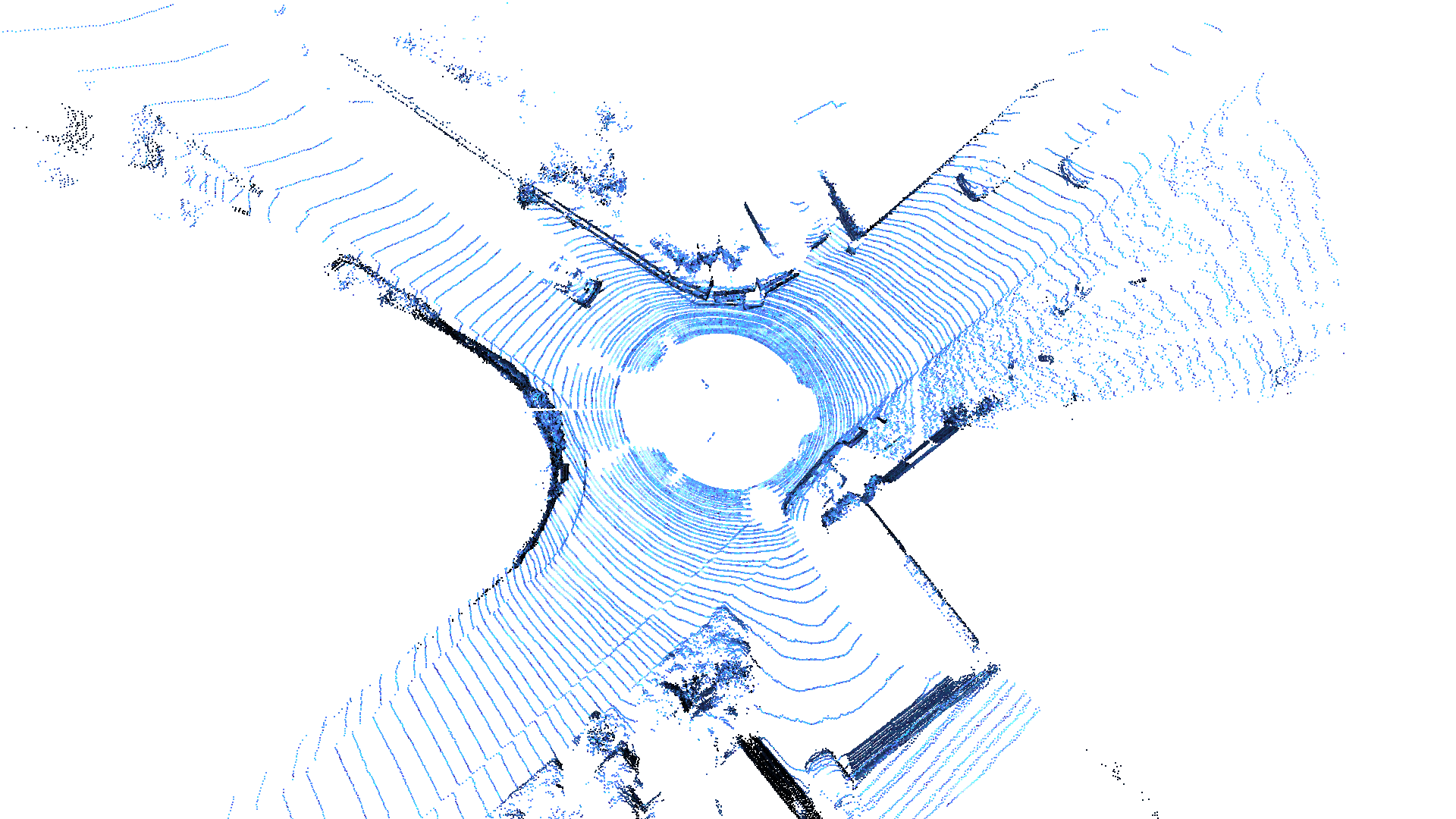}} \\
  \vspace{2em}
  \includegraphics[width=1.0\linewidth]{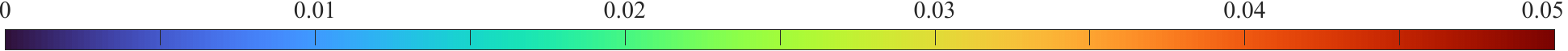}
  \caption{Visualization of reconstruction quality of sample ``11\_000000.bin'' at different quantization precisions.}
  \label{Fig:AppendixVis}
\end{figure*}

To further evaluate the effectiveness of the proposed method, we visualize compression results at three different bitrates, ranging from low to high. As shown in Fig.~\ref{Fig:AppendixVis}, the proposed method consistently produces reconstructed point clouds with lower distortion at comparable bitrates. These visual results align well with the findings in the Rate-Distortion Performance section, further supporting its advantage in compression performance.


